\documentclass[draft]{agujournal2019}

\usepackage{url} %this package should fix any errors with URLs in refs.
\usepackage{lineno}
\usepackage[inline]{trackchanges} %for better track changes. finalnew option will compile document with changes incorporated.
\usepackage{soul}
\usepackage{amsmath}
\usepackage[ruled,vlined]{algorithm2e}
\usepackage{setspace}

\graphicspath{ {./figures/} }

\draftfalse

\journalname{Water Resources Research}

\begin{document}

%\title{Subsurface Characterization using Generative Models: why ensemble-based approaches perform better than naive optimization-based approaches}
\title{Subsurface Characterization using Ensemble-based Approaches with Deep Generative Models}
\authors{Jichao Bao\affil{1}, Hongkyu Yoon\affil{2}, and Jonghyun Lee\affil{1}}

\affiliation{1}{University of Hawaii at Manoa, Honolulu, HI, USA}
\affiliation{2}{Sandia National Laboratories, Albuquerque, NM, USA}

\correspondingauthor{Jonghyun Lee}{jonghyun.harry.lee@hawaii.edu}

\begin{keypoints}
\item Subsurface characterization using the Wasserstein generative adversarial network with gradient penalty.
\item Gaussian, channelized, and fractured fields are tested to demonstrate the accuracy and efficiency of our approach.
\item The ensemble-based and optimization-based approaches are compared to demonstrate why the ensemble-based approach performs better with GANs.
\end{keypoints}

\begin{abstract}
Estimating spatially distributed properties such as hydraulic conductivity ($K$) from available sparse measurements is a great challenge in subsurface characterization. However, the use of inverse modeling is limited for ill-posed, high-dimensional applications due to computational costs and poor prediction accuracy with sparse datasets. In this paper, we combine Wasserstein Generative Adversarial Network with Gradient Penalty (WGAN-GP), a deep generative model that can accurately capture complex subsurface structure, and Ensemble Smoother with Multiple Data Assimilation (ES-MDA), an ensemble-based inversion method, for accurate and accelerated subsurface characterization. WGAN-GP are trained to generate high-dimensional $K$ fields from a low-dimensional latent space and ES-MDA then updates the latent variables by assimilating available measurements. Several subsurface examples are used to evaluate the accuracy and efficiency of the proposed method and the main features of the unknown $K$ fields are characterized accurately with reliable uncertainty quantification. 

Furthermore, the estimation performance is compared with a widely-used variational, i.e., optimization-based, inversion approach, and the proposed approach outperforms the variational inversion method, especially for the channelized and fractured field examples. We explain such superior performance by visualizing the objective function in the latent space: because of nonlinear and aggressive dimension reduction via generative modeling, the objective function surface becomes extremely complex while the ensemble approximation can smooth out the multi-modal surface during the minimization. This suggests that the ensemble-based approach works well over the variational approach when combined with deep generative models at the cost of forward model runs unless convergence-ensuring modifications are implemented in the variational inversion.
\end{abstract}

\section{Introduction}
Subsurface characterization is critical for subsurface energy storage and recovery processes and risk management associated with subsurface resevoir activities \cite{tartakovsky2013assessment,newell2017investigation,kang2020improved,kang2021hydrogeophysical,ghorbanidehno2020recent,yoon2013parameter}. However, heterogeneous aquifer properties such as hydraulic conductivity is a great challenge for subsurface characterization because of the limited number of observations and uncertainty in numerical models. Data assimilation (DA) approach offers a solution to integrate dynamic data (e.g., time-series pressure) into numerical models, identify unknown model parameters, and reduce the estimation uncertainty \cite{liu2007uncertainty, liu2012advancing}. Among various DA methods, Kalman-type methods are widely used for their optimality and efficiency for linear-Gaussian problems while code-intrusive implementation of the adjoint state method and successive linearizations are required for nonlinear, non-Guassian applications \cite{carrassi2018data}. The ensemble Kalman filter (EnKF) \cite{evensen1994sequential} is one of the most popular Kalman-type methods for its simplicity and easy implementation without intrusive code changes. For example, \citeA{chen2006data} applied EnKF to continuously update hydraulic conductivity and hydraulic head by assimilating dynamic and static data. The results demonstrated that the estimated conductivity field using EnKF matches the reference field very well. Moreover, incorrect prior knowledge can be rectified to some extent through DA procedure. \citeA{liu2008investigation} implemented EnKF to investigate the flow and transport processes at the macro-dispersion experiment (MADE) site in Columbus, Mississippi, USA. The advection-dispersion (AD) model and the dual-domain mass transfer (DDMT) model were used to analyze the tritium plume. Piezometric head measurements and tritium concentrations were assimilated to estimate the hydraulic conductivity and major parameters of the AD and DDMT models such as dispersion coefficients and mass transfer rates. This work demonstrated that EnKF is an efficient method for solving large-scale, nonlinear fluid flow and transport problems. \citeA{li2012jointly} applied EnKF to jointly estimate hydraulic conductivity and porosity by assimilating hydraulic head and concentration data. The results also showed that integrating more and different types of data could improve predictions of groundwater flow and solute transport.

One disadvantage of EnKF is that the update process requires repeated parameter estimation and restarting the simulation at each time step, which is time-consuming and difficult to implement when multiphysics models are involved \cite{zhang2018iterative}. Various methods have been proposed to address the challenges, and ensemble smoother (ES) \cite{van1996data} is an alternative to EnKF. All the data are simultaneously assimilated through ES, which significantly reduces the computational burden. However, the data are used only once through ES, i.e., ensemble Kalman update is performed only once in each iteration, to achieve a global update, which might not be able to produce acceptable results due to premature convergence to the optimal estimate. Therefore, some iterative methods have been proposed \cite{gu2007iterative, chen2012ensemble}. Ensemble smoother with multiple data assimilation (ES-MDA) \cite{emerick2013ensemble} is one of the most promising approaches. The observed data can be used multiple times by ES-MDA and iteratively integrated into the models, i.e., Kalman update is corrected through iterative linearizations in each iteration, to achieve a better result. \citeA{tavakoli2013comparison} and \citeA{emerick2016analysis} investigated the performance of ES-MDA in history matching of production and seismic data as well as CO2 saturation in geologic carbon storage. The results showed that ES-MDA can produce plausible estimates of reservoir properties with a reasonable match to the observed production data. \citeA{fokker2016application} implemented ES-MDA to estimate the subsurface model parameters of the Bergermeer gas field. The line-of-sight measurements were used in the DA process and ES-MDA yielded reasonable estimates of the compaction coefficient and elastic modulus. 

The Kalman-type, which is considered as linear(ized)-Gaussian, methods share a common problem, i.e., the limitation of Gaussian assumption \cite{ghorbanidehno2020recent}. These Gaussian prior-based methods can provide optimal solutions if the prior follows a multi-Gaussian distribution, but there are also many non-Gaussian situations such as channelized aquifers that these methods cannot handle \cite{zhou2014inverse}. Although a variety of approaches such as power transformation from non-Gaussian to Gaussian prior or Monte Carlo sampling have been proposed to deal with the non-Gaussian cases \cite<e.g.,>[]{oliver1997markov, hu2000gradual, le2015history}, the heavy computational burden may hinder the application of these methods. Moreover, DA and inverse models even with the Gaussian prior are prohibited by high-dimensional problems with expensive forward models unless one considers fast linear algebra \cite{wang2021pbbfmm3d}, dimension reduction \cite{lee2014large,ghorbanidehno2020recent}, and/or reduced order modeling \cite{kadeethum2021framework}. In recent decades, the fast growth of deep learning (DL) and its impressive applications in many fields provide a new direction to address these non-Gaussian estimations and computationally challenging issues in the DA. Specifically, deep generative models \cite{goodfellow2014generative} have attracted significant attention due to their promising ability to represent data distributions and generate new samples in an efficient and accurate fashion. Deep generative models can be used for different tasks such as generating synthetic but realistic images, enhancing image resolution, and recovering missing parts of data \cite{turhan2018recent}.

Among many deep generative models, generative adversarial network (GAN) \cite{goodfellow2014generative} has become one of the widely used generative models in DA because of its capability to sample from the target distribution. GAN includes a generator and a discriminator. The generator converts randomly sampled latent variables into generated data, and the discriminator determines whether the data are authentic \cite{grover2018flow}. Several attempts to apply GAN in the subsurface systems have been conducted. \citeA{sun2018discovering} presented a state-parameter identification GAN (SPID-GAN) to learn the bidirectional mappings between the high-dimensional parameter space and the corresponding model state space. Groundwater flow modeling was conducted and the results demonstrated that SPID-GAN performed well in approximating the bidirectional state-parameter mappings. \citeA{laloy2018training} introduced a spatial GAN (SGAN) to generate high-dimensional complex media samples. 2D steady-state flow and 3D transient hydraulic tomography cases were presented to show the performance of SGAN-based inversion approach. The results illustrated that the SGAN-based approach can produce earth model realizations that are similar to the true model and fit the data well. \citeA{janssens2020computed} applied GAN to improve the resolution of computed tomography (CT) scans. The results showed that the GAN-based super-resolution method can better characterize the pore networks and fluid flow properties. Using the improved super-resolution CT scans as input resulted in more accurate simulations. \citeA{patel2021gan} used GAN-based prior to solve Bayesian inverse problems in a Monte Carlo framework and one of their applications is to estimate the permeability field. 

Despite the impressive applications, the drawbacks of n\=aive GAN are mode collapse and training instability. Mode collapse is the generator collapsing to generate only one or a small subset of different outputs or modes. The Hessian of the loss function in a GAN is indefinite, so the optimal solution is therefore to find a saddle point instead of a local minimum. The stochastic gradient methods commonly used for training cannot reliably converge to a saddle point, and converging to a saddle point requires good initialization, which leads to training instability \cite{creswell2018generative}. To address these issues, variants of GAN have been proposed \cite<e.g.,>[]{nowozin2016f, mao2017least, zhao2016energy, berthelot2017began}. \citeA{arjovsky2017wasserstein} introduced the Wasserstein GAN (WGAN) with a new metric called earth-mover (or Wasserstein-1) distance to measure the distance between the generated distribution and real distribution. The weight clipping strategy was used in the original WGAN implementation. However, \citeA{gulrajani2017improved} found that simply clipping the weights might limit the ability of the discriminator, and they provided another solution called gradient penalty. This new variant of GAN was named WGAN-GP, in which the norm of discriminator gradients with respect to data samples was penalized during training. The results showed that WGAN-GP could stabilize the training process and alleviate the mode collapse problem \cite{kadeethum2022continuous}. Furthermore, it is expected that the gradient of WGAN-GP is more regular than that of ``vanilla'' GAN so that the data assimilation techniques based on gradient evaluation of the objective function would benefit and converge to the optimal solution with a smaller number of iterations.

While WGAN-GP has shown better performance in complex image generation, it has rarely been used for subsurface DA or inverse problems. In this paper, we will propose a WGAN-GP and ES-MDA based subsurface characterization approach and apply it to several inverse modeling examples (Gaussian, channelized, and fractured aquifers) to show the performance of our proposed method. Based on our applications, we will show the highly non-linear objective function surface of the data assimilation problem when used with GAN, which is a side product of the aggressive data dimension reduction via GAN, and experimentally show that ensemble-based methods would be better performed than the variational, i.e., optimization-based, inverse modeling approach at the cost of more forward model runs, which will be our additional contribution of this paper. The rest of this paper is organized as follows: the methodology is introduced in Section~\ref{sec:method}. Synthetic applications are shown in Section~\ref{sec:applications}, and some discussions on the benefit of ensemble-based approaches are presented in Section~\ref{sec:discussion}. Finally, conclusions are summarized in Section~\ref{sec:conclusion}.

\section{Methodology}
\label{sec:method}
\subsection{Wasserstein Generative Adversarial Network with Gradient Penalty}
\subsubsection{Background}
The training strategy of GAN can be defined as a minimax game between two competing networks, i.e., a generator network and a discriminator network \cite{goodfellow2014generative}:
\begin{equation}
    \mathop{\min}_{G}\mathop{\max}_{D} E_{x \sim P_{x}}[logD(x)]+E_{\mathbf{z} \sim P_{z}}[log(1-D(G(\mathbf{z})))]
\end{equation}
where $x$ represents the training data and $P_{x}$ is the data distribution; $G(\mathbf{z})$ represents the generated samples ($G(\mathbf{z})= \widetilde{x} \sim P_{g}$, and $P_{g}$ is the generator distribution); $\mathbf{z}$ represents the latent space variables sampled from a simple distribution $P_{z}$ such as a Gaussian distribution; $G$ is the generator network that maps the input variables $\mathbf{z}$ to the data space, and $D$ is the discriminator network that evaluates the probability that the data comes from the data distribution $P_{x}$ or the generator distribution $P_{g}$. During training, the discriminator attempts to assign correct labels to both training data and samples from $P_{g}$ while the generator tries to fool the discriminator. Through this interactive process, the generator $G$ learns a distribution $P_{g}$ that is close to the data distribution $P_{x}$.

However, commonly used metrics in GAN such as Kullback-Leibler (KL) divergence and Jensen-Shannon (JS) \cite{goodfellow2014generative} may not be continuous with respect to the generator's parameters, which makes training difficult. Therefore, \citeA{arjovsky2017wasserstein} proposed to use the so-called earth-mover (or Wasserstein-1) distance $W(P_{x},P_{g})$ as an alternative:
\begin{equation}
    W(P_{x},P_{g}) = \mathop{inf}_{\gamma \in \prod(P_{x},P_{g})} E_{(x,y) \sim \gamma}	\left[ \left \|x-y  \right\|  \right]
\end{equation}
where $\prod(P_{x},P_{g})$ represents all joint distributions $\gamma (x,y)$ whose marginal distributions are $P_{x}$ and $P_{g}$ respectively. The earth-mover (EM) distance is the minimum cost of moving ``mass" from $x$ to $y$ in order to transform the distribution $P_{x}$ into $P_{g}$.

\begin{figure}[htbp]
\centering
\includegraphics[scale=0.2, angle=0]{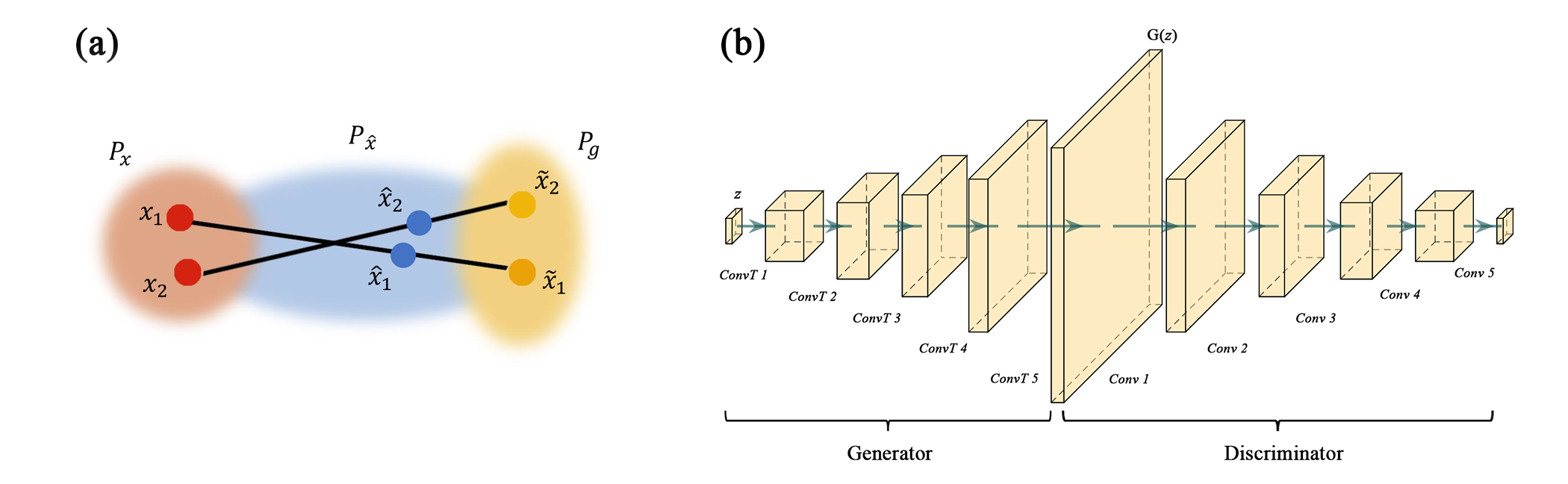}
\caption{(a) Schematic diagram of gradient penalty. The red area represents the data distribution $P_{x}$, the yellow area represents the generator distribution $P_{g}$, and the blue area represents the penalty distribution $P_{\hat{x}}$. $\hat{x_{i}} \sim P_{\hat{x}}$ indicates random samples from the straight lines connecting $x_{i}$ and $\widetilde{x_{i}}$ ($x_{i} \sim P_{x}$, and $\widetilde{x_{i}} \sim P_{g}$). (b) The architecture of WGAN-GP. $\mathbf{z}$ represents the latent space variables, $G(\mathbf{z})$ indicates the images generated by the generator $G$, $ConvT$ is the transposed convolutional layer, and $Conv$ represents the convolutional layer.
}
\label{fig:WGAN_GP}
\end{figure}

Under mild assumptions, the EM distance $W(P_{x},P_{g})$ is continuous everywhere and differentiable almost everywhere \cite{arjovsky2017wasserstein}. The loss function in Wasserstein GAN (WGAN) is defined using the Kantorovich-Rubinstein duality \cite{villani2009optimal} as follows:
\begin{equation}
    \mathop{\min}_{G}\mathop{\max}_{D \in \mathcal{D}} E_{x \sim P_{x}}[D(x)]+E_{\mathbf{z} \sim P_{z}}[D(G(\mathbf{z}))]
\end{equation}
where $\mathcal{D}$ is a set of 1-Lipschitz functions, i.e., discriminator whose rate of change is bounded by a constant (=1) \cite{o2006metric}. The discriminator $D$ in WGAN is called a ``critic'' since it is a real-valued function rather than a classifier, but we will still use the word ``discriminator'' to keep it consistent with the existing GAN structure. \citeA{arjovsky2017wasserstein} proposed a weight clipping strategy to constrain the weights in a compact space [-c,c] (e.g., [-0.01,0.01]) to enforce the Lipschitz continuity. However, the weight clipping might limit the networks' capabilities and cause gradient vanishing or exploding during training. \citeA{gulrajani2017improved} introduced an alternative way to enforce the Lipschitz constraint called the gradient penalty in which the discriminator loss is defined as:
 \begin{equation}
    L_D = \underbrace{ E_{\mathbf{z} \sim P_{z}}[D(G(\mathbf{z}))] - E_{x \sim P_{x}}[D(x)] }_{\text{original discriminator loss}} + \lambda\underbrace{  E_{\hat{x} \sim P_{\hat{x}}}[(\left \| \nabla _{\hat{x}}D(\hat{x})  \right\|_2-1)^2] }_{\text{gradient penalty}}
\end{equation}
where the random samples $\hat{x} \sim P_{\hat{x}}$ land on the straight lines between pairs of points sampled from the data distribution $P_{x}$ and the generator distribution $P_{g}$ (Figure \ref{fig:WGAN_GP} (a)). Enforcing the gradient constraint everywhere is intractable, but enforcing it only along the straight lines would be sufficient for the Lipschitz condition and can lead to good training performance. The factor $\lambda$ is set to 10 as suggested in their paper \cite{gulrajani2017improved}.

\subsubsection{Training}
\begin{figure}[htbp]
\centering
\includegraphics[scale=0.2, angle=0]{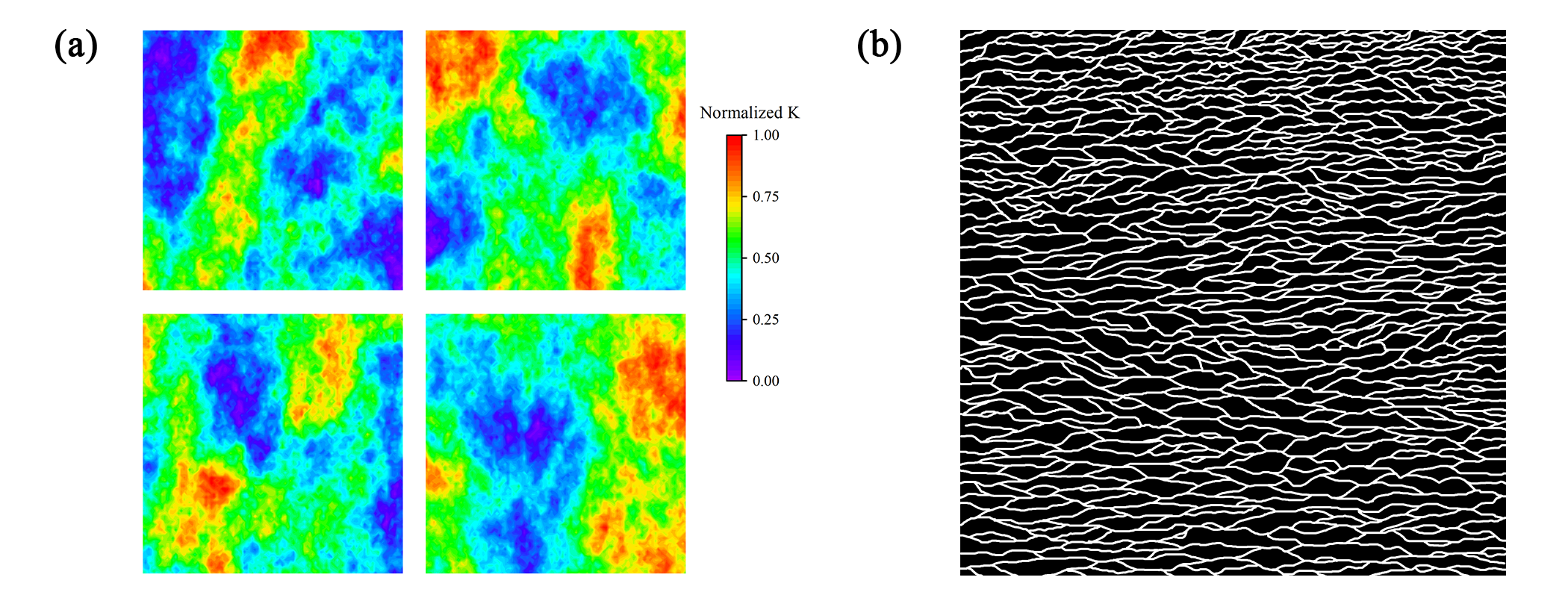}
\caption{(a) Examples of Gaussian training images. (b) Channelized subsurface training image (2500$\times$2500) from \cite{laloy2018training}.}
\label{fig:Training_Images}
\end{figure}

The architecture of WGAN with gradient penalty (WGAN-GP) used in this paper is illustrated in Figure \ref{fig:WGAN_GP} (b). Two training sessions were performed using a Gaussian (log-normal) hydraulic conductivity field dataset shown in Figure~\ref{fig:Training_Images} (a) and a channelized binary field dataset shown in Figure~\ref{fig:Training_Images} (b). The trained generators were then applied to synthetic data assimilation experiments. Each dataset contains 80,000 images and the size of each image is 96 $\times$ 96. Gaussian training images were generated by the sequential Gaussian simulation \cite{gomez1993joint} using SGeMS software~\cite{remy2009applied}. The images were produced using simple Kriging with a spherical variogram. The nugget effect is 0 and the contribution is 1. The search ellipsoid has ranges of 60 and 40 with 0 azimuth. The values were normalized to [0, 1] for the training purpose as in Figure~\ref{fig:Training_Images} (a). For channelized aquifer characterization case, 80,000 training images (size 96 $\times$ 96) were randomly chosen from the 2500 x 2500 multi-point geostatistical training image \cite{laloy2018training} as shown in Figure~\ref{fig:Training_Images} (b).

The details of network parameters for the Gaussian field dataset are shown in Table \ref{tab:WGAN_Gaussian}. In the table, $k$ represents the kernel size, $s$ represents the stride, $p$ represents the zero-paddings, $d$ represents the dilation, and InsNorm2d represents the instance normalization. The generator was trained with two-dimensional transposed convolutional layer (ConvT2d). The activation function of the first 3 layers was ReLU, and the final layer was a Sigmoid function.  The discriminator was trained using 2D convolutional layers (Conv2d). The activation functions of the first 3 layers were LeakyReLU(0.2), and there was no activation function for the final layer. The latent space $z$ was set to $6 \times 6$. For the channelized dataset the network parameters are shown in Table~\ref{tab:WGAN_Channel}. With hyperparameter optimization adding one more layer to the generator and the discriminator of the deep learning model for the Gaussian field resulted in a better result for the channelized data with the latent space dimension $z$ set to $3 \times 3$. All models were constructed using PyTorch \cite{paszke2019pytorch}. 

\begin{table}[htbp]\small
\renewcommand\arraystretch{1.5}
 \caption{\label{tab:WGAN_Gaussian}The generator and discriminator architectures for Gaussian data}
 \centering
 \begin{tabular}{c|c}
 \hline
  Generator & Discriminator \\
 \hline
 Input: $6 \times 6$ latent space $\mathbf{z}$ \hspace{3.3cm} & Input: $96 \times 96$ image \hspace{4.75cm} \\
 Layer 1: ConvT2d, 4k 2s 1p 1d, InsNorm2d, ReLU & Layer 1: Conv2d, 4k 2s 1p 1d, InsNorm2d, LeakyReLU  \\
 Layer 2: ConvT2d, 4k 2s 1p 1d, InsNorm2d, ReLU & Layer 2: Conv2d, 4k 2s 1p 1d, InsNorm2d, LeakyReLU  \\
 Layer 3: ConvT2d, 4k 2s 1p 1d, InsNorm2d, ReLU & Layer 3: Conv2d, 4k 2s 1p 1d, InsNorm2d, LeakyReLU  \\
 Layer 4: ConvT2d, 4k 2s 1p 1d, Sigmoid \hspace{1.35cm} & Layer 4: Conv2d, 4k 2s 1p 1d \hspace{3.5cm} \\
 Output: $96 \times 96$ image \hspace{3.85cm} & Output: $6 \times 6$ array \hspace{4.85cm}  \\
 \hline
 \end{tabular}
\end{table}

\begin{table}[htbp]\small
\renewcommand\arraystretch{1.5}
 \caption{\label{tab:WGAN_Channel}The generator and discriminator architectures for channelized data}
 \centering
 \begin{tabular}{c|c}
 \hline
  Generator & Discriminator \\
 \hline
 Input: $3 \times 3$ latent space $\mathbf{z}$ \hspace{3.3cm} & Input: $96 \times 96$ image \hspace{4.75cm} \\
 Layer 1: ConvT2d, 4k 2s 1p 1d, InsNorm2d, ReLU & Layer 1: Conv2d, 4k 2s 1p 1d, InsNorm2d, LeakyReLU  \\
 Layer 2: ConvT2d, 4k 2s 1p 1d, InsNorm2d, ReLU & Layer 2: Conv2d, 4k 2s 1p 1d, InsNorm2d, LeakyReLU  \\
 Layer 3: ConvT2d, 4k 2s 1p 1d, InsNorm2d, ReLU & Layer 3: Conv2d, 4k 2s 1p 1d, InsNorm2d, LeakyReLU  \\
 Layer 4: ConvT2d, 4k 2s 1p 1d, InsNorm2d, ReLU & Layer 4: Conv2d, 4k 2s 1p 1d, InsNorm2d, LeakyReLU  \\
 Layer 5: ConvT2d, 4k 2s 1p 1d, Sigmoid \hspace{1.35cm} & Layer 5: Conv2d, 4k 2s 1p 1d \hspace{3.5cm} \\
 Output: $96 \times 96$ image \hspace{3.85cm} & Output: $3 \times 3$ array \hspace{4.85cm}  \\
 \hline
 \end{tabular}
\end{table}

The training process was conducted using an NVIDIA TITAN V GPU card. The training for the Gaussian dataset took 2.5 hours with 60 epochs. The training for the channelized dataset took 3.3 hours with 50 epochs. The batch size and learning rate were set to 32 and 1e-4, respectively. 

\begin{figure}[htbp]
\centering
\includegraphics[scale=0.3, angle=0]{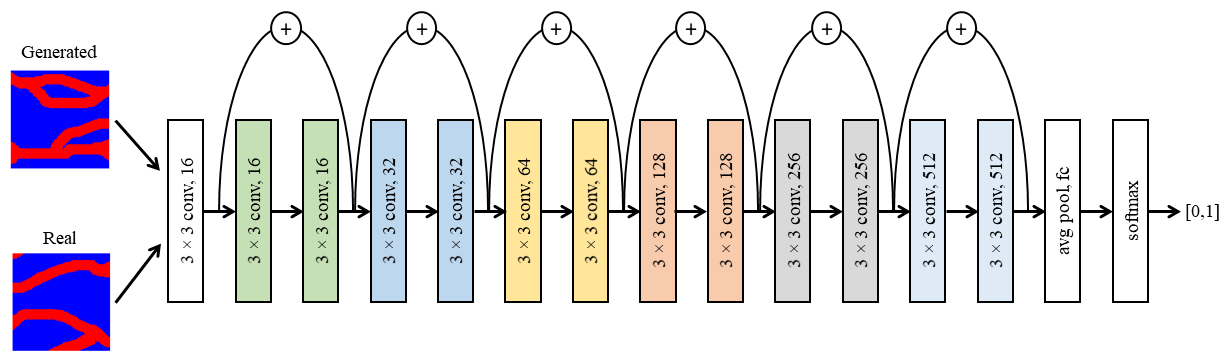}
\caption{The architecture of ResNet. The residual blocks are marked with different colors. ``3×3 conv, 16'' represents the convolutional layer with 3×3 kernel and the output has 16 channels. ``avg pool" represents the average pooling layer, ``fc'' represents the fully connected layer, and ``softmax'' represents the softmax activation function.
}
\label{fig:ResNet}
\end{figure}

Residual networks (ResNets) \cite{he2016deep} were used as the classifier/discriminator and they were trained with the real samples from Figure~\ref{fig:Training_Images} and samples generated by the pre-trained generators. The real samples were labeled as ``1" and the generated samples were labeled as ``0". The architecture of the ResNet-based discriminator is shown in Figure \ref{fig:ResNet}. To evaluate the performance of the trained generator, we performed the nearest neighbor sample test \cite{lopez2016revisiting, xu2018empirical} where the generated samples were classified with the same label as their nearest neighbors. If the generator distribution is the same as the data distribution ($P_{g}=P_{r}$), generated samples should have an equal probability of being classified as real data or generated data, then classification becomes like the random guess. Consequently, the nearest neighbor sample test should yield around $50\%$ accuracy when $P_{g}=P_{r}$. The details about the nearest neighbor sample tests and efficient calculation can be found in \cite{xu2018empirical}. The nearest neighbor sample tests were performed by using 2,000 unlabeled images from the training datasets and 2,000 unlabeled images generated by the trained generators stored over the entire training epochs.

\begin{figure}[htbp]
\centering
\includegraphics[scale=0.14, angle=0]{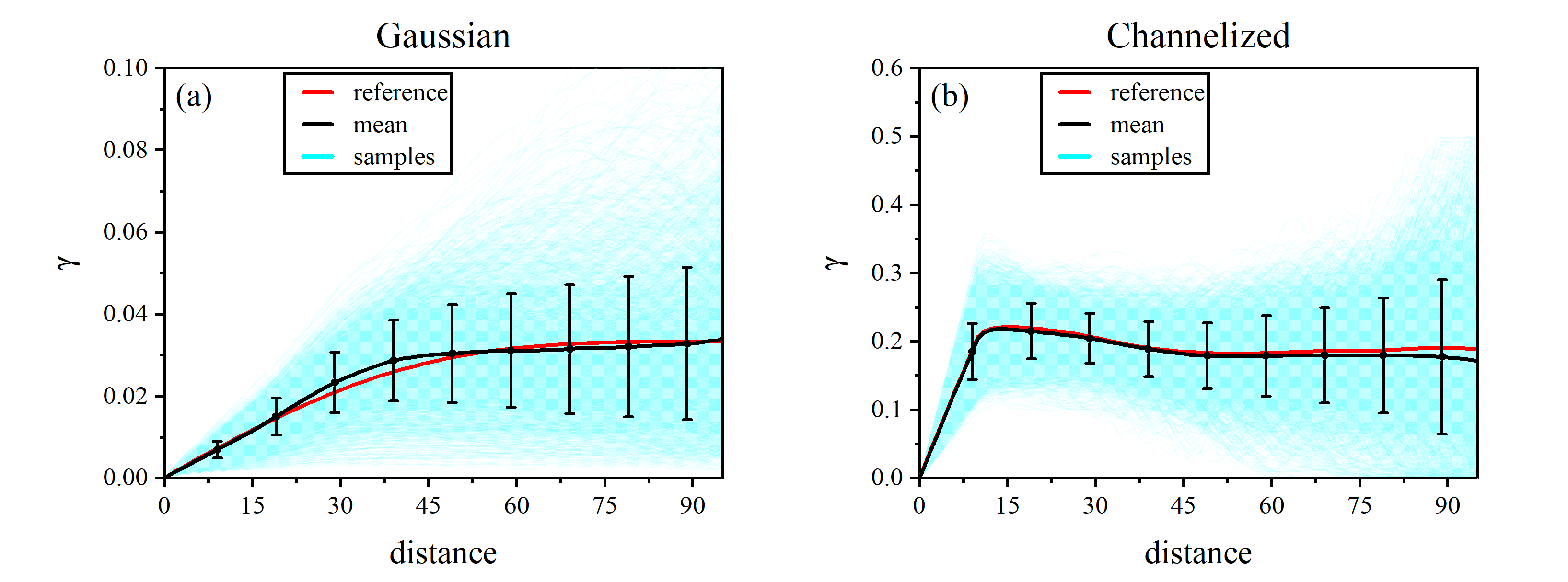}
\caption{(a) The semivariograms for the Gaussian data;  (b) The semivariograms for the channelized data. The red line indicates the reference semivariogram computed from the training images. The blue lines represent the semivariograms of 2,000 generated images. The black line represents the mean, and the error bars indicate the standard deviation of the generated 2,000 samples. }
\label{fig:Variograms}
\end{figure}

\begin{figure}[htbp]
\centering
\includegraphics[scale=0.1, angle=0]{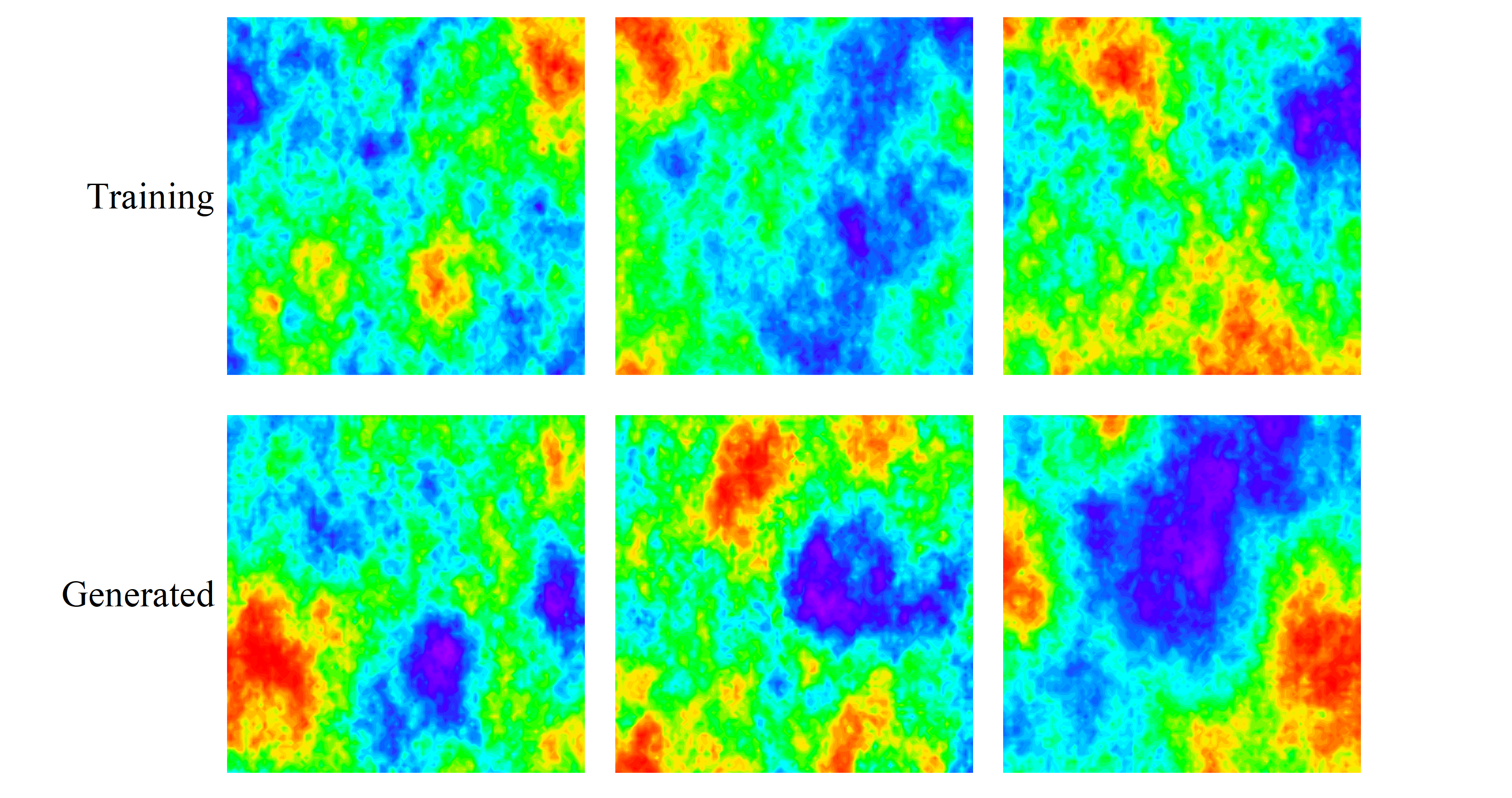}
\caption{Examples of Gaussian training images (top) and generated images  by WGAN-GP (bottom)}
\label{fig:Relz_Gaussian}
\end{figure}

\begin{figure}[htbp]
\centering
\includegraphics[scale=0.1, angle=0]{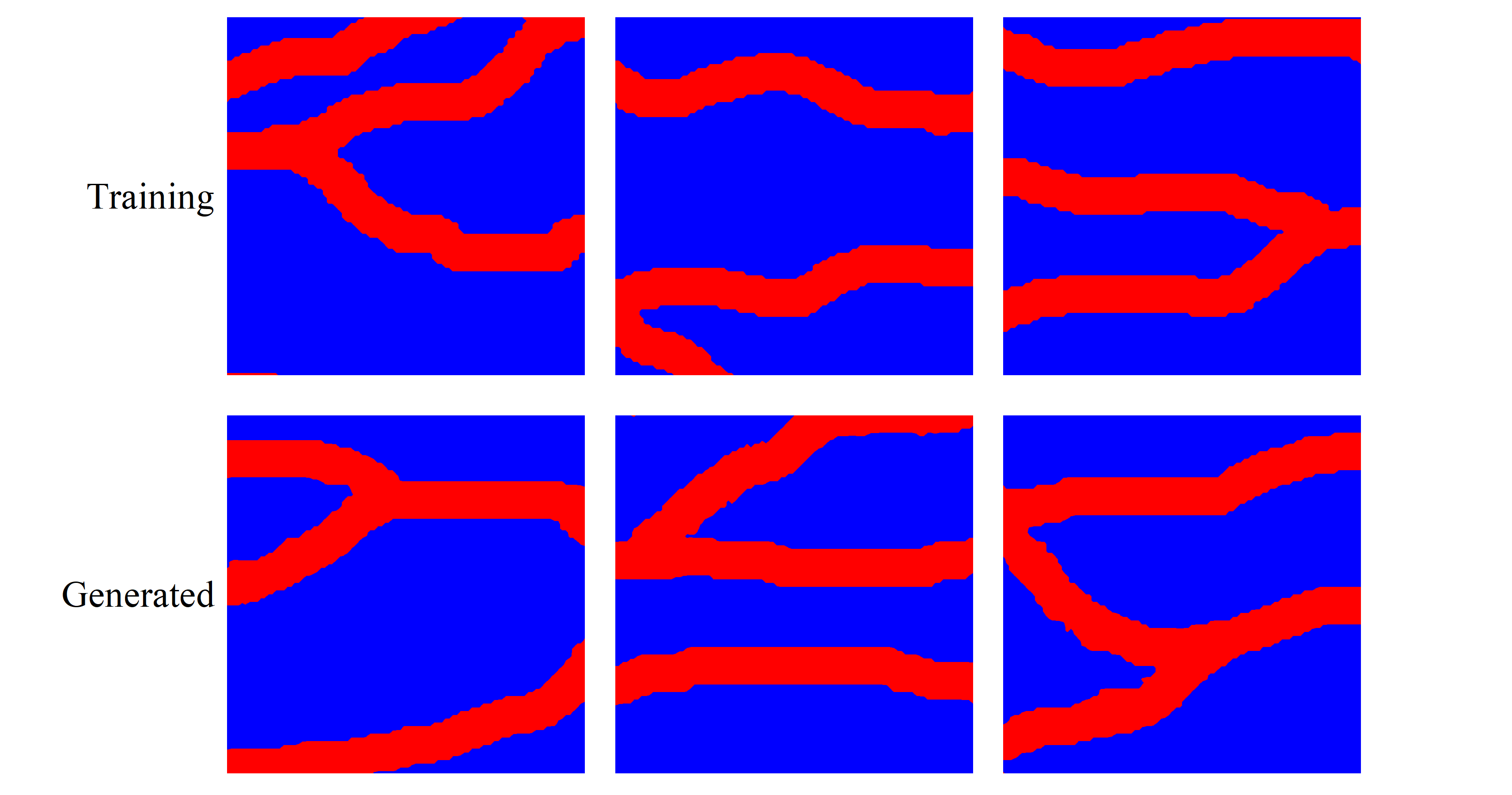}
\caption{Examples of channelized training images (top) and generated images by WGAN-GP (bottom)}
\label{fig:Relz_Channel}
\end{figure}

The accuracy reached a minimum value close to $50\%$ after 55 and 40 epochs of training with the Gaussian dataset and the channelized dataset, respectively. The results became stable after that. Therefore, the trained generators of epoch 55 and epoch 40 were chosen to perform the following experiments. The semivariogram along the west-east direction was used to show the generative ability of the trained generators:

\begin{equation}
    \gamma(h) = \frac{1}{2N(h)} \sum _{i=1}^{N(h)} \left[ f(u_{i}+h)-f(u_{i})) \right]^2
\end{equation}
where $h$ is the separation distance, $N(h)$ is the number of pairs of locations, and $f(u_{i}+h)$ and $f(u_{i})$ are the values at point $u_{i}+h$ and point $u_{i}$, respectively. Figure \ref{fig:Variograms} (a) shows the semivariograms after 55 epochs of training for the Gaussian dataset. 2,000 images generated by the trained generator were used to calculate the semivariograms. The mean of semivariograms of the original 2,000 images from the training dataset was used as the reference (red in Figure \ref{fig:Variograms} (a)). The mean of the generated samples is pretty close to the reference. Figure \ref{fig:Variograms} (b) shows the semivariograms for the channelized dataset after 40 epochs of training. The mean of the generated samples matches the reference well. Figure \ref{fig:Relz_Gaussian} and Figure \ref{fig:Relz_Channel} are randomly chosen training and generated images that illustrate the trained generators can produce realistic images similar to the training images.

\subsection{Ensemble Smoother with Multiple Data Assimilation}
Here we follow ES-MDA \cite{emerick2013ensemble} to update the latent space variables $\mathbf{z}$. We will revisit our choice of ES-MDA for general inverse and data assimilation problems and discuss justifications later. The parameter matrix at the $i$-th step $\mathbf{Z^{i}}$ can be expressed as follows:
\begin{equation}
    \mathbf{Z}^{i} = \begin{bmatrix} \mathbf{z}_1^{i} & \mathbf{z}_2^{i} & \cdots & \mathbf{z}_{N_r}^{i}\end{bmatrix} = \begin{bmatrix}
    z_{11}^{i} & z_{12}^{i} & \cdots & z_{1N_r}^{i}\\
    z_{21}^{i} & z_{22}^{i} & \cdots & z_{2N_r}^{i}\\
    \vdots & \vdots & \ddots & \vdots\\
    z_{N_d1}^{i} & z_{N_d2}^{i} & \cdots & z_{N_dN_r}^{i}\\
    \end{bmatrix}
\end{equation}
where $\mathbf{z}_j^{i}$ is $j$-th realization of the latent space variable vector whose $k$-th element $\mathbf{z}_{jk}$ is $j$-th latent variable in $k$-th realization where $j = 1, \cdots, N_d$ and $k = 1, \cdots, N_r$. $N_d$ and $N_r$ represent the number of latent variables and realizations, respectively. In inverse modeling or batch-data assimilation, the forward model can be defined as the following form:
\begin{equation}
    \mathbf{d}=F(\mathbf{k})+\mathbf{\epsilon}
\end{equation}
where $\mathbf{d}$ is the $N_{d} \times 1$ data vector, such as the hydraulic head at observation locations, $F(.)$ is the forward operator, such as USGS Groundwater Flow Model MODFLOW-2005 \cite{harbaugh2005modflow}, $\mathbf{k}$ is the $N_{k} \times 1$ model parameter vector, and $\mathbf{\epsilon}$ is the model error. 
The objective is to estimate the unknown model parameters that reproduce the observation data $\mathbf{d}_{obs}$, and the updating process in ES-MDA follows the Kalman update through ensemble covariance matrix approximation\cite{evensen1994sequential}:
\begin{equation}
    \label{eq:ensemble_kalman_update}
    \mathbf{z}_{j}^{i+1}=\mathbf{z}_{j}^{i}+\mathbf{C}_{ZD}^{i}(\mathbf{C}_{DD}^{i} +\alpha_{i}\mathbf{C}_{D})^{-1}(\mathbf{d}_{uc,j}^{i}-\mathbf{d}_{j}^{i}), \;j=1, \cdots , N_{r}
\end{equation}
where $\mathbf{C}_{ZD}^{i}$ is the cross-covariance between the latent variables and the simulated data, $\mathbf{C}_{DD}^{i}$ is the auto-covariance of the simulated data, $\mathbf{C}_{D}$ is the observation error covariance, and $\mathbf{d}_{uc}$ is the perturbed observation data defined as:
\begin{equation}
    \mathbf{d}_{uc}=\mathbf{d}_{obs}+\sqrt{\alpha_{i}}\mathbf{C}_{D}^{1/2}\epsilon_d
\end{equation}
where $\alpha_{i}$ is the inflation coefficient of the $i$-th iteration ($\sum_{i=1}^{N_{a}}\frac{1}{\alpha_{i}}=1$, $N_{a}$ is the number of iterations) and $\epsilon_{d}\sim N(0,\mathbf{I}_{N_d})$. 

In the updating equation~\ref{eq:ensemble_kalman_update}, the inverse of the auto-covariance matrix $\mathbf{C}_{DD}^{i}+\alpha_{i}\mathbf{C}_{D}$ needs to be computed. However, the matrix may be singular due to the rank $N_r$ ensemble covariance approximation and its small eigenvalues may cause numerical instability and error, thus truncated eigenvalue decomposition is applied to obtain the pseudo-inverse of the auto-covariance matrix, and the measurement error covariance $\mathbf{C}_{D}$ is rescaled with the Cholesky decomposition $\mathbf{C}_{D} = \mathbf{C}_{D}^{1/2}(\mathbf{C}_{D}^{1/2})^{\top}$ \cite{emerick2012history}. With the eigenvalue truncation, we obtain
\begin{equation}
    \mathbf{C}_{DD}^{i}+\alpha_{i}\mathbf{C}_{D} \approx \mathbf{U_{n}}\mathbf{\Lambda_{n}}\mathbf{V_{n}}^{T}
\end{equation}
where $\mathbf{\Lambda_{n}}$ is a diagonal matrix containing $N_{n}$ largest eigenvalues. The truncation threshold parameter $N_{n}$ is defined by the following truncation scheme:
\begin{equation}
    \frac{\sum_{i=1}^{N_{n}} \lambda_{i}}{\sum_{i=1}^{N_{t}} \lambda_{i}} \leq E
\end{equation}
where $\lambda_{i}$ is the $i$-th singular values sorted in decreasing order, $N_{t}$ is the total number of singular values, and $E$ is the energy of the eigenvalues retained, typically between 0.9 and 1.0. Thus, $N_{n}$ is the number that makes the ratio of the sum of the $N_{n}$ largest singular values to the sum of the total singular values less than or equal to $E$. The inverse of matrix $\mathbf{C}_{DD}^{i}+\alpha_{i}\mathbf{C}_{D}$ can then be approximated as follows:
\begin{equation}
    \left(\mathbf{C}_{DD}^{i}+\alpha_{i}\mathbf{C}_{D}\right)^{-1} \approx \mathbf{V_{n}}\mathbf{\Lambda_{n}}^{-1}\mathbf{U_{n}}^{T}
\end{equation}

\subsection{Coupling WGAN-GP with ES-MDA}
\begin{figure}[htbp]
\centering
\includegraphics[scale=0.3, angle=0]{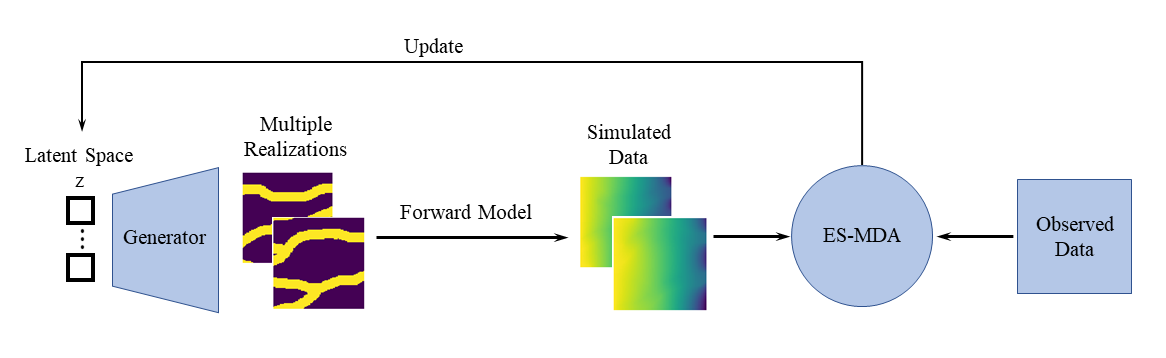}
\caption{The flowchart of coupling WGAN-GP with ES-MDA.}
\label{fig:WGAN_ESMDA}
\end{figure}

Based on the previous developments, we couple WGAN-GP within the ES-MDA framework to characterize spatially distributed hydraulic conductivity fields from sparse datasets. The use of WGAN-GP will encode the data-driven prior distribution and accelerate the estimation of unknown spatial parameters in the encoded, smaller latent space by reducing the computational costs associated with matrix-matrix multiplications, e.g., Jacobian-covariance products ~\cite{ghorbanidehno2020recent}. The procedure of coupling WGAN-GP with ES-MDA is shown in Figure \ref{fig:WGAN_ESMDA}. The trained generator is used to map the latent space $\mathbf{z}$ to multiple realizations of the hydraulic conductivity fields from the prior distribution. By simply running ``black-box'' forward models, the ensemble-based approaches only need to obtain the simulated outputs such as hydraulic heads without the code-intrusive adjoint-state formulation that explicitly constructs Jacobian and its products. ES-MDA updates the latent variables iteratively based on the mismatch between simulated data and observed data until the convergence as shown in Algorithm \ref{AL:Coupling}.

\begin{algorithm}[H]
\setstretch{1.2}
\caption{Coupling WGAN-GP with ES-MDA}
\DontPrintSemicolon
    $\text{Set: }N_{a} = \text{The number of iterations}$\;
    $\text{Set: }\mathbf{d}_{obs} = \text{Observed data (e.g., hydraulic head)}$\;
    $\text{Set: }N_{r} = \text{The number of realizations}$\;
\Begin{
    \text{Sample initial $\mathbf{z}$ from a Gaussian distribution $N(0,1)$}\;
    \For{$i = 1, 2, \cdots, N_{a}$}{
        $\alpha_{i}=N_{a}$\;
        \For{$j = 1, 2, \cdots, N_{r}$}{
            \text{Generate the hydraulic conductivity fields with the trained generator: } $\mathbf{K}_{j}^{i} = G(\mathbf{z}_{j}^{i})$\;
            \text{Run forward model to obtain the hydraulic head data } $\mathbf{d}_{j}^{i}$\;
            \text{Perturb the observation data: } $\mathbf{d}_{uc,j}^{i}=\mathbf{d}_{obs}+\sqrt{\alpha_{i}}\mathbf{C}_{D}^{1/2}\mathbf{I}_d$\;
        }
    \text{Calculate: } $\mathbf{C}_{ZD}^{i} = \frac{1}{N_{r}-1}\sum_{j=1}^{N_{r}}(\mathbf{z}_{j}^{i}-\bar{\mathbf{z}^{i}})
            (\mathbf{d}_{j}^{i}-\bar{\mathbf{d}^{i}})^\mathrm{T}$\;
    \text{Calculate: } $\mathbf{C}_{DD}^{i} = \frac{1}{N_{r}-1}\sum_{j=1}^{N_{r}}(\mathbf{d}_{j}^{i}-\bar{\mathbf{d}^{i}})
            (\mathbf{d}_{j}^{i}-\bar{\mathbf{d}^{i}})^\mathrm{T}$\;
    \text{Update: } $\mathbf{z}_{j}^{i+1}=\mathbf{z}_{j}^{i}+\mathbf{C}_{ZD}^{i}(\mathbf{C}_{DD}^{i}
            +\alpha_{i}\mathbf{C}_{D})^{-1}(\mathbf{d}_{uc,j}^{i}-\mathbf{d}_{j}^{i}), j=1, ..., N_{r}$\;
    }
\textbf{end}
}
\label{AL:Coupling}
\end{algorithm}

\section{Applications}
\label{sec:applications}
In this section, two benchmark applications demonstrate the performance of our proposed inversion approach. In the first example, we test with a Gaussian field, where the log-transformed hydraulic conductivity follows a Gaussian distribution. Note that this example can be readily addressed by the traditional inverse modeling methods and their linear dimension reduction variants \cite[e.g.,]{kitanidis1995quasi,lee2014large}. The second example is a channelized, binary field with two categories: low conductivity areas with high conductivity channels. These benchmark fields are log-transformed and normalized in [-1, 1] for the training purpose. The true fields for Gaussian and channelized fields are randomly selected, neither used for training nor generated by trained generators.
\subsection{Gaussian Fields}
\subsubsection{Model Setup}
\begin{figure}[htbp]
\centering
\includegraphics[scale=0.2, angle=0]{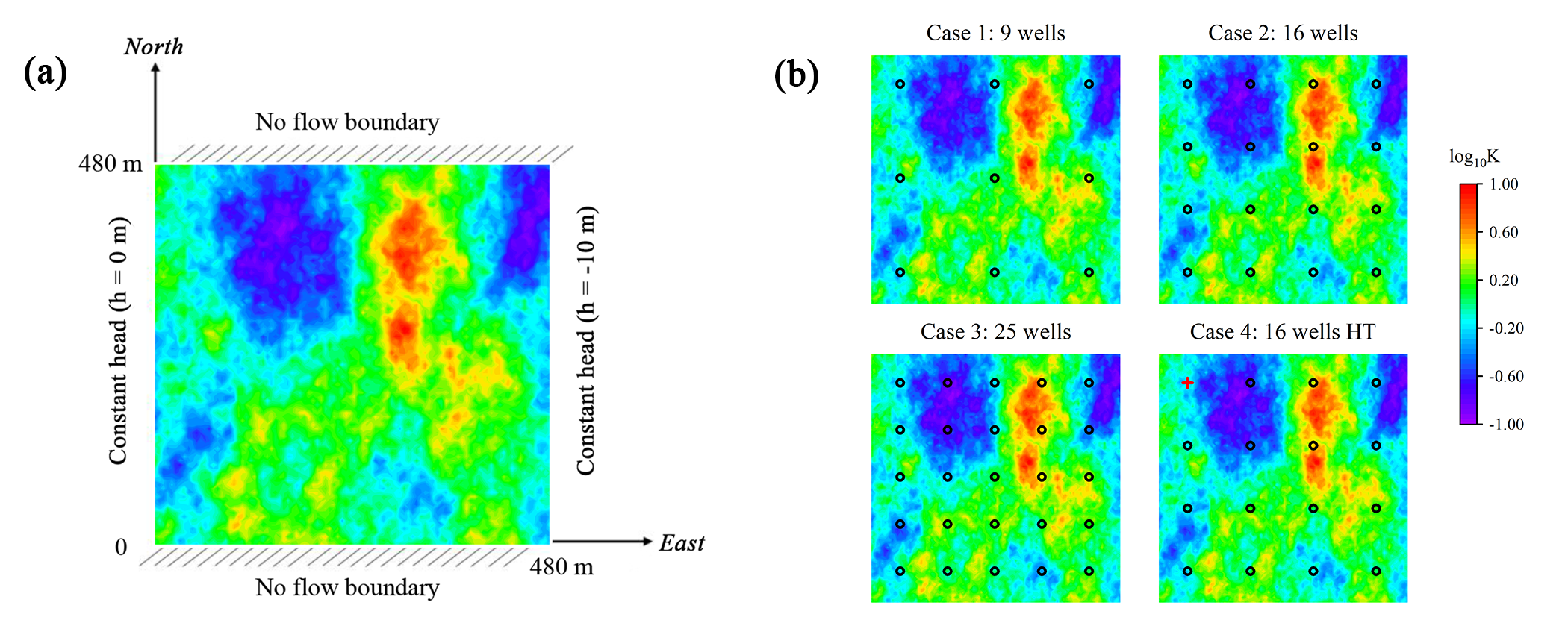}
\caption{(a) Model settings. No flow on northern and southern boundaries, and constant heads on the western and eastern boundaries are imposed. (b) Well locations for Cases 1-4. Case 4 indicates the hydraulic tomography (HT), and aquifer characterization through multiple cross-well pumping tests. and the red cross represents an example of the pumping location and it moves to the next location once the pumping test is completed. The black circles indicate the locations of monitoring wells.}
\label{fig:Model}
\end{figure}

The true field and model settings are shown in Figure \ref{fig:Model} (a). A confined aquifer (480 m $\times$ 480 m $\times$ 10 m) is discretized into 96 rows $\times$ 96 columns $\times$ 1 layer. The logarithmic hydraulic conductivity ranges from -1 to 1 (0.1 m/d to 10 m/d). The northern and southern boundaries are set to no flow boundaries. The western and eastern sides are set to the constant head boundaries with h = 0 m and -10 m, respectively. The recharge rate is set to 0.001 m/d. The model is constructed using the USGS groundwater flow model MODFLOW-2005 \cite{harbaugh2005modflow} with its Python interface FloPy \cite{bakker2016scripting}, and steady-state groundwater flow is simulated for our demonstrations.

\subsubsection{Results with Different Observation Layouts}
Four cases with different data availability were tested to evaluate the performance of different observation strategies. Figure~\ref{fig:Model} (b) shows the true field with well locations. Cases 1-3 have 9, 16, and 25 monitoring wells, respectively. The hydraulic tomography (HT) in which we perform cross-well experiments \cite{gottlieb1995identification,yeh2000hydraulic,lee2014large} is applied in Case 4. Case 4 shows an example of a single pumping experiment; the red cross indicates the pumping well location with the pumping rate of 50 m$^3$/d and the black circles represent monitoring wells so that each pumping test produces 15 steady-state head measurements. Once a pumping test is completed, the pumping well moves to the next location and continues the experiment, which results in 16 (pumping tests) $\times$ 15 (measurements/test) = 240 measurements in total. A Gaussian error of 0.02 is added to the simulated measurements representing forward modeling and data collection errors. 

\begin{figure}[htbp]
\centering
\includegraphics[scale=0.4, angle=0]{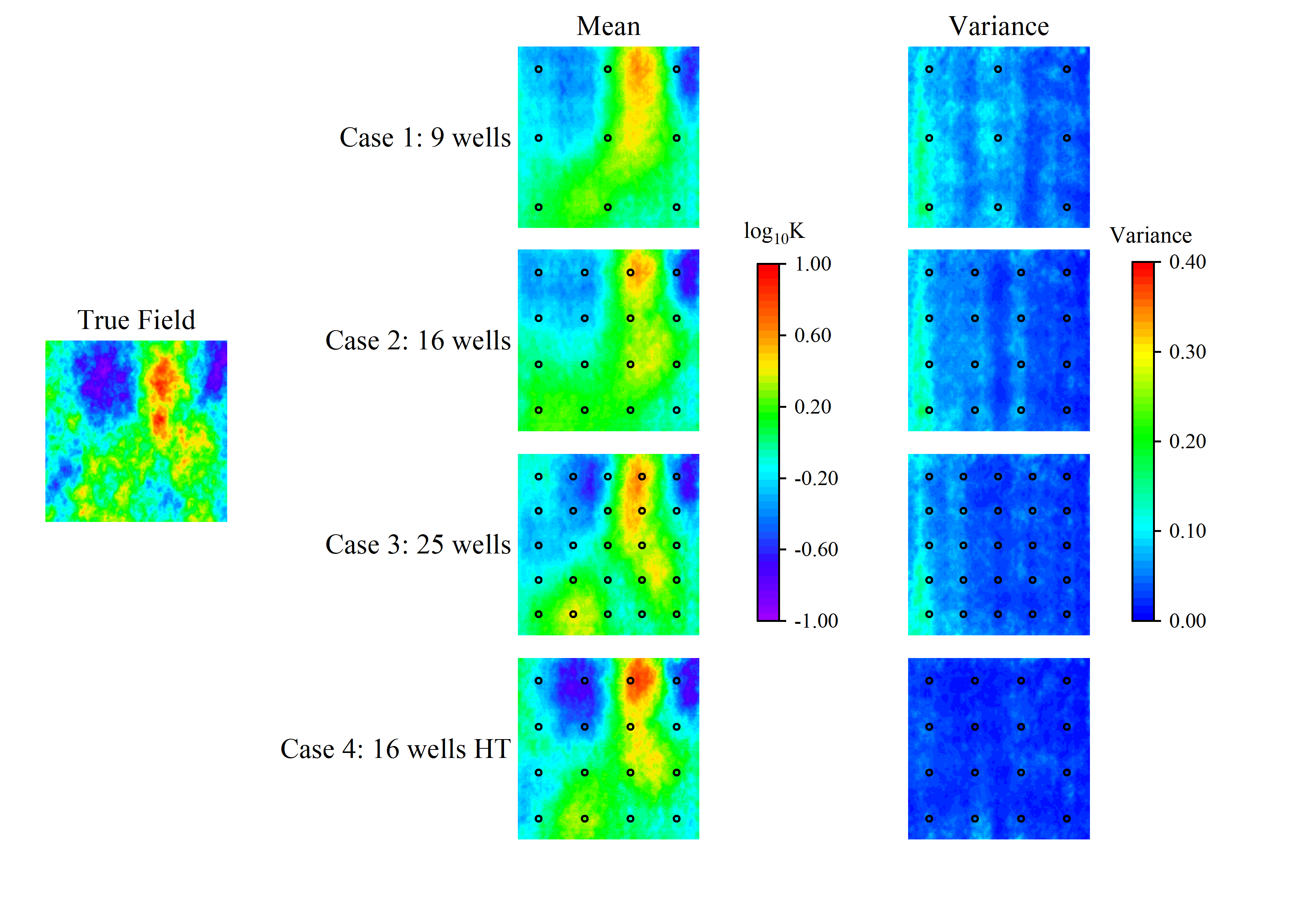}
\caption{True Field (left), estimated mean (middle), and posterior variance (right) for Gaussian Benchmark Cases. The black circles indicate well locations.
}
\label{fig:MeanVar_Gaussian}
\end{figure}

\begin{figure}[htbp]
\centering
\includegraphics[scale=0.3, angle=0]{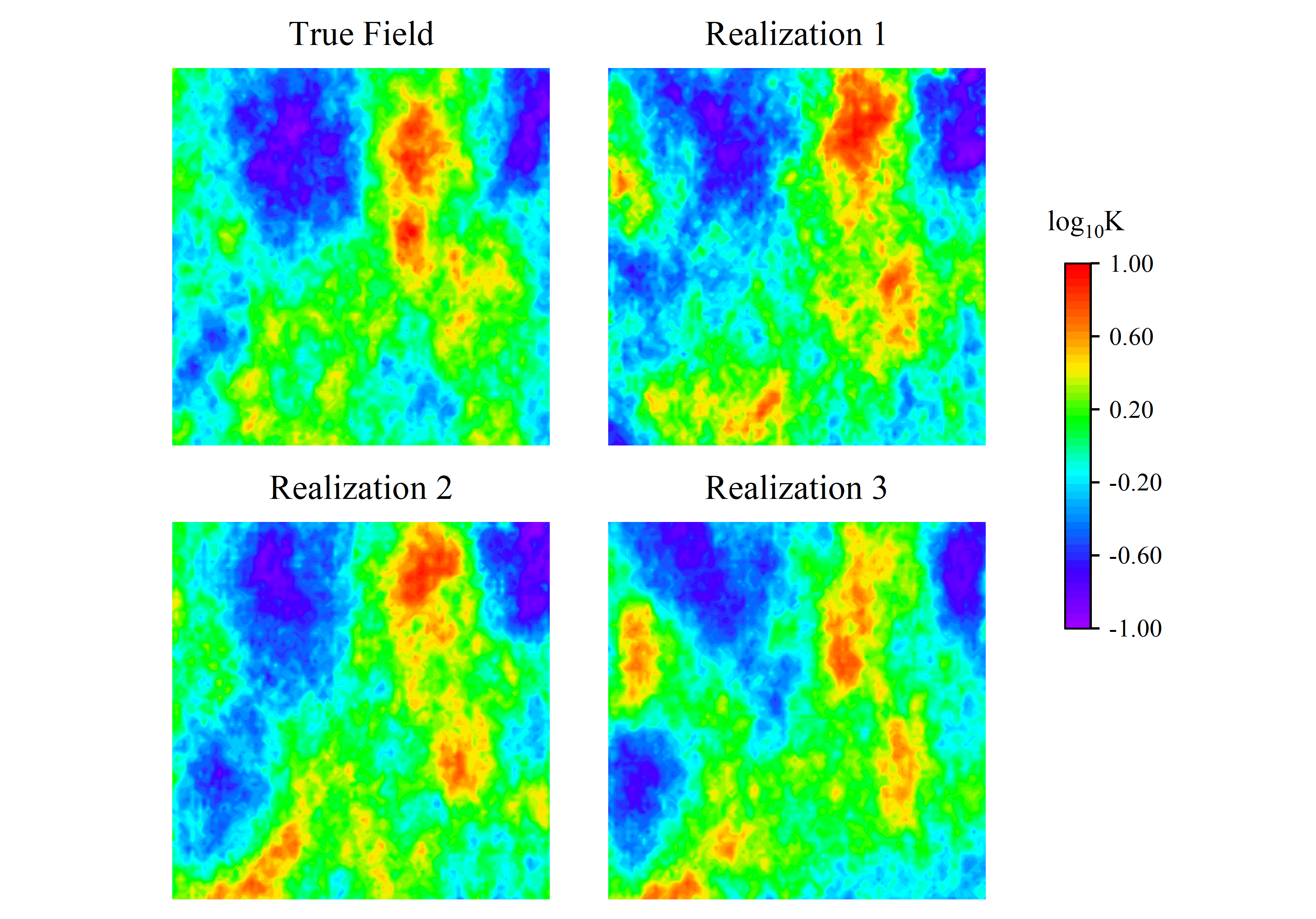}
\caption{True Gaussian field (top left) and three realizations from the posterior pdf in Case 4.
}
\label{fig:Realizations}
\end{figure}

\begin{figure}[htbp]
\centering
\includegraphics[scale=0.2, angle=0]{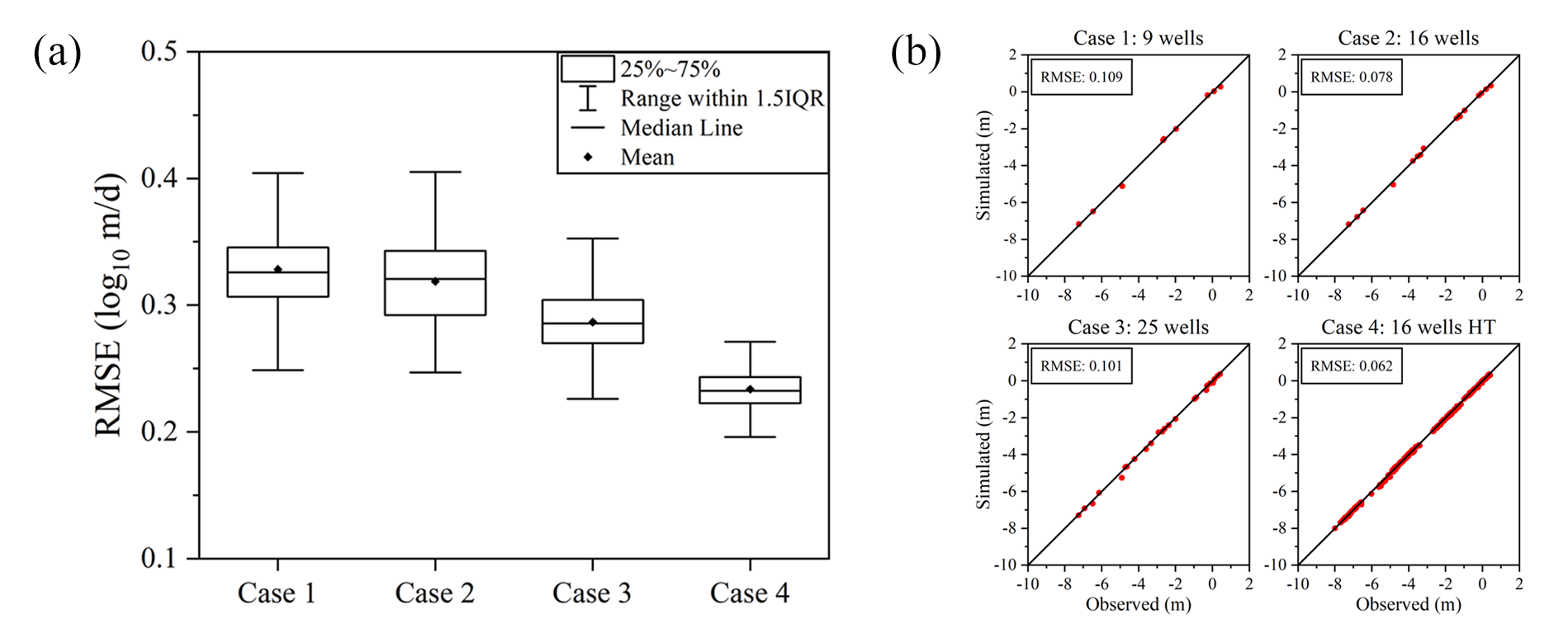}
\caption{(a) Boxplot of RMSE (log$_{10}\ $m/d) of estimated hydraulic conductivity for Gaussian cases. (b) simulated hydraulic heads vs. observed hydraulic heads for Gaussian cases. The black lines are $45^{\circ}$ lines.}
\label{fig:Fit_Gaussian}
\end{figure}

After hyperparameter calibrations \cite{kitanidis1991orthonormal}, the number of ensembles is set to 200 with the number of maximum iterations to 8. The mean and variance of 200 realizations for different cases are shown in Figure~\ref{fig:MeanVar_Gaussian}. For Cases 1-3, the mean is closer to the true field as more wells are available. Placing more monitoring wells is a straightforward way to improve results at the expense of the well installation and experiment budgets. With HT, we obtain better results with fewer wells as shown in Figure~\ref{fig:MeanVar_Gaussian} where the result of Case 4 is much better than those from the other cases (see the accuracy reported in Figure~\ref{fig:Fit_Gaussian}). We also evaluate the uncertainty of the estimate using the variance computed from the posterior realizations and it is observed that HT reduces uncertainty significantly as expected. Figure~\ref{fig:Realizations} shows 3 realizations of Case 4. The proposed approach can reconstruct the main features in the true field, especially with HT.

Figure \ref{fig:Fit_Gaussian} (a) is a boxplot showing the Root Mean Square Error (RMSE) of the hydraulic conductivity of 200 realizations in each case. RMSE is calculated as follows:

\begin{equation}
    RMSE = \sqrt{\frac{1}{N_{k}}\sum_{i=1}^{N_{k}}(\mathbf{K}_{i}-\mathbf{T}_{i})^{2}}
\end{equation}
where $\mathbf{K}_{i}$ indicates the estimated parameter value and $\mathbf{T}_{i}$ is the true value, $N_{k}$ is the total number of parameters.

It is shown that the mean of the RMSE decreases as more wells are used as expected. The RMSE using HT in Case 4 is much smaller than the other three cases. The spread in Case 4 is also the smallest, indicating less uncertainty as well as less variability within the conditional realizations. For this Gaussian case, principal component analysis or variational autoencoder-based inversion approaches will work with a similar performance \cite{lee2014large,forghani2022variational}. 

The fitting error of the simulated hydraulic head and the observed hydraulic head is shown in Figure~\ref{fig:Fit_Gaussian} (b). The red dots indicate the mismatch between observed hydraulic heads and the simulated hydraulic heads, which is the simulation result using estimated mean hydraulic conductivity. The RMSE values are also displayed as an indicator of fitting error in Figure~\ref{fig:Fit_Gaussian} (b). Case 4 has the smallest value, which demonstrates that the inversion results using HT are more accurate as expected.

\subsection{Channelized Fields}
In this subsection, we tested our proposed method to estimate the subsurface field with high permeable channels, which is more challenging than the previous Gaussian case and the traditional methods would fail to find an accurate solution because of the Gaussian prior assumption \cite{kadeethum2021framework,kang2022integration}. Inversion performance was evaluated at various pumping rates and with measurement error levels. 
\subsubsection{Different Pumping Rates}
\begin{figure}[htbp]
\centering
\includegraphics[scale=0.4, angle=0]{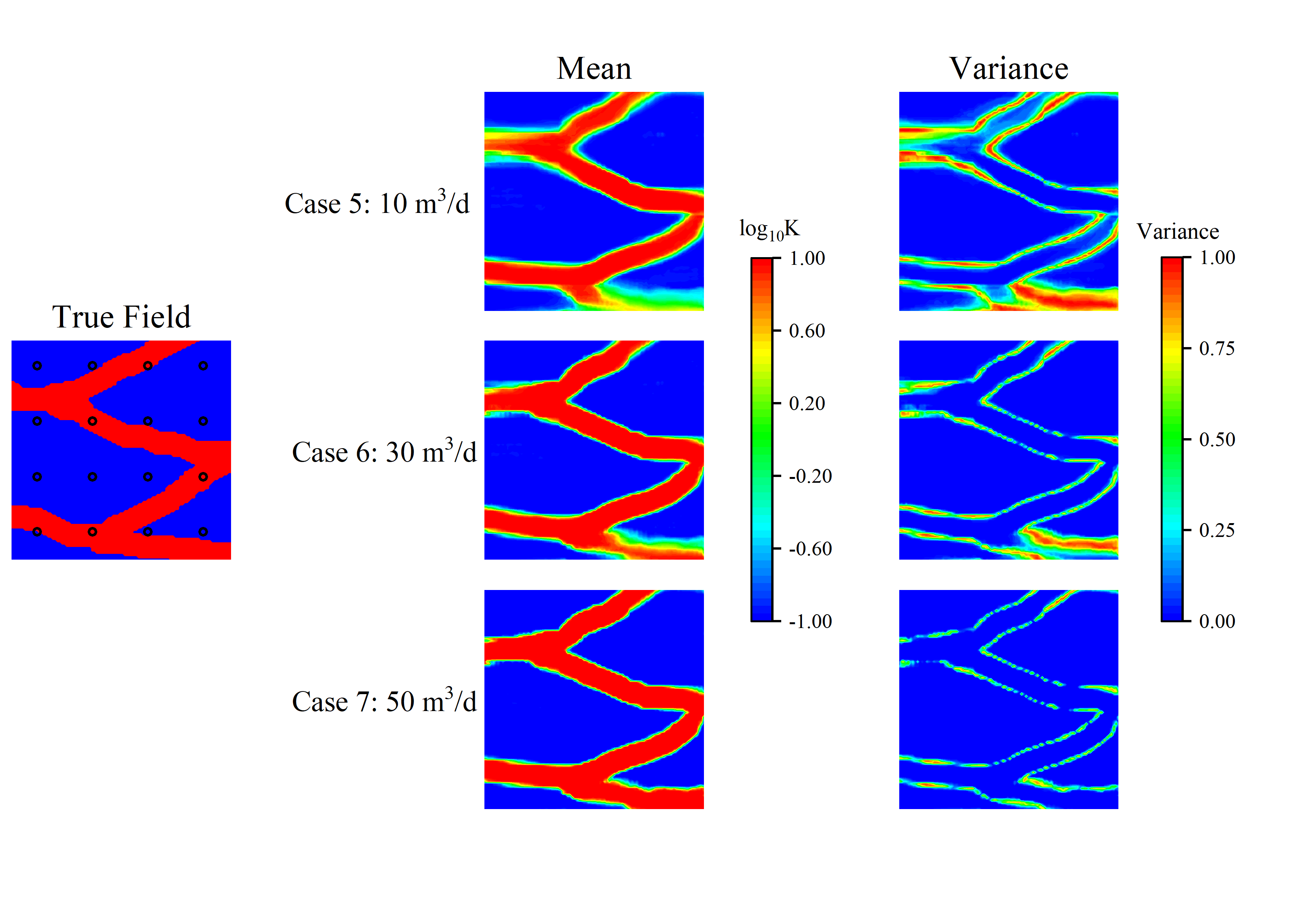}
\caption{Mean and variance for the channelized field. The true field is shown on the left, and the black circles indicate well locations.
}
\label{fig:MeanVar_Channel}
\end{figure}

The true field is shown on the left in Figure \ref{fig:MeanVar_Channel}. The red channels represent connected high conductivity materials such as sand or volcanic lava tubes (K = 10 m/d), and the blue areas represent low conductivity materials such as clay (K = 0.1 m/d). The black circles indicate well locations. Three cases HT with different pumping rates (10 m$^3$/d, 30 m$^3$/d, and 50 m$^3$/d) were evaluated. A Gaussian error of 0.02 is added to the simulated measurements.

\begin{figure}[htbp]
\centering
\includegraphics[scale=0.2, angle=0]{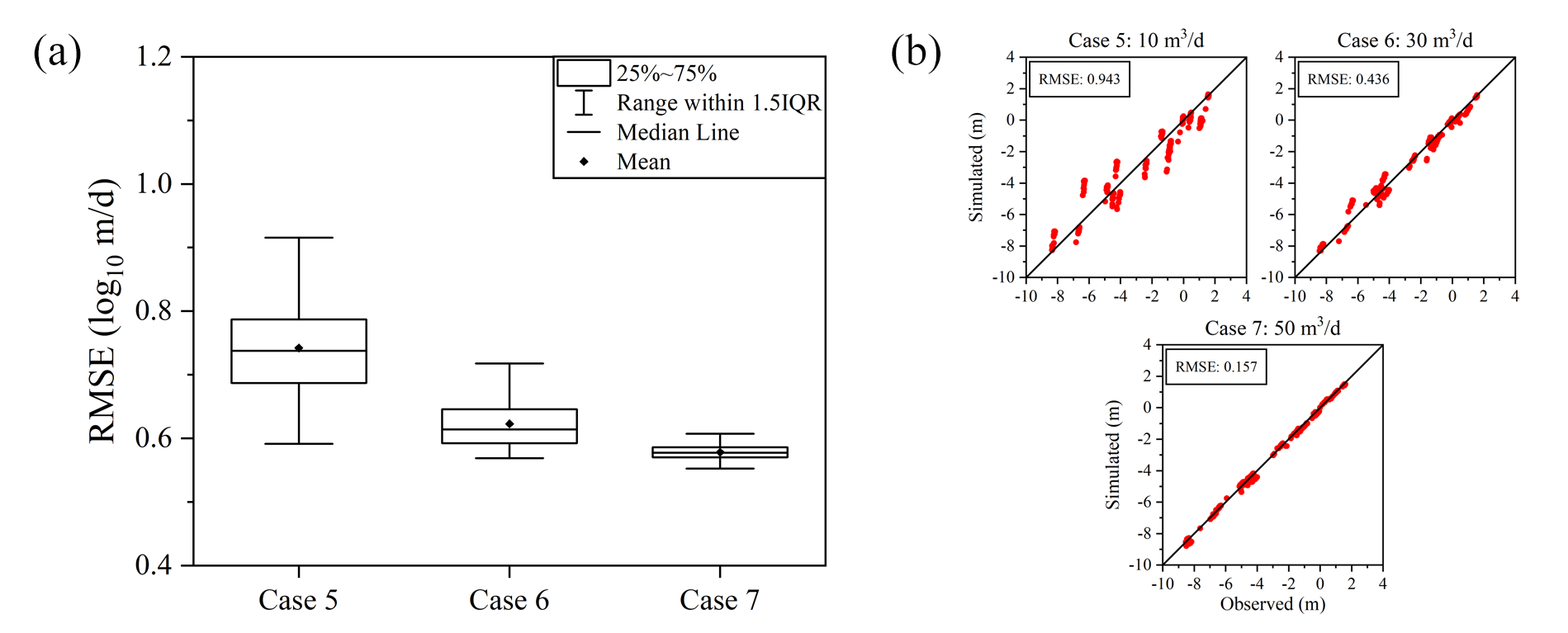}
\caption{(a) Boxplot of hydraulic conductivity RMSE (log$_{10}\ $m/d) and (b) simulated hydraulic heads vs. observed hydraulic heads for the channelized $K$ cases. The black lines are $45^{\circ}$ lines.}
\label{fig:Fit_Channel}
\end{figure}

ES-MDA with the same configuration (200 realizations and 8 iterations) is applied to the channelized field characterization. The results are shown in Figure~\ref{fig:MeanVar_Channel} illustrating that data collected at a larger pumping rate leads to an improved result. The mean of 200 realizations becomes closer to the true field as the pumping rate increases. The posterior variance plots reflect (linearized) estimation uncertainty with larger uncertainty around the channel boundaries as expected in the previous research \cite[e.g.]{lee2013bayesian} and the variance of the result with 50 m$^3$/d pumping rate is lower than the other two cases. Figure \ref{fig:Fit_Channel} (a) also shows the RMSE of estimated hydraulic conductivity for each case confirming the RMSE value decreases dramatically and the uncertainty is significantly reduced as the pumping rate increases. This is expected since the sensitivity of the data, i.e., Jacobian of the forward model $\frac{\partial F}{\partial \mathbf{k}}$, increases as the pumping rate increases, thus the measurement information that contains important features becomes less contaminated by the measurement error~\cite{kitanidis1998observations}. Still, the inversion with data at the small pumping rate can identify the structure of underlying channels by incorporating the meaningful data-driven prior from WGAN-GP in the inversion. 

The fitting errors with different pumping rates are shown in Figure~\ref{fig:Fit_Channel} (b). Like in the Gaussian test cases, the red dots in Figure~\ref{fig:Fit_Channel} (b) display the mismatch between the observed hydraulic heads and the simulated hydraulic heads using estimated mean hydraulic conductivity. Note that the mean RMSE values from the smaller pumping rate experiments are larger than the simulated error level of 0.02 because of high variability and uncertainty in the final ensemble that originates from inaccurate estimation around the high permeable channel boundaries as shown in Figure~\ref{fig:MeanVar_Channel}.

\subsubsection{Different Error Levels}
\begin{figure}[htbp]
\centering
\includegraphics[scale=0.4, angle=0]{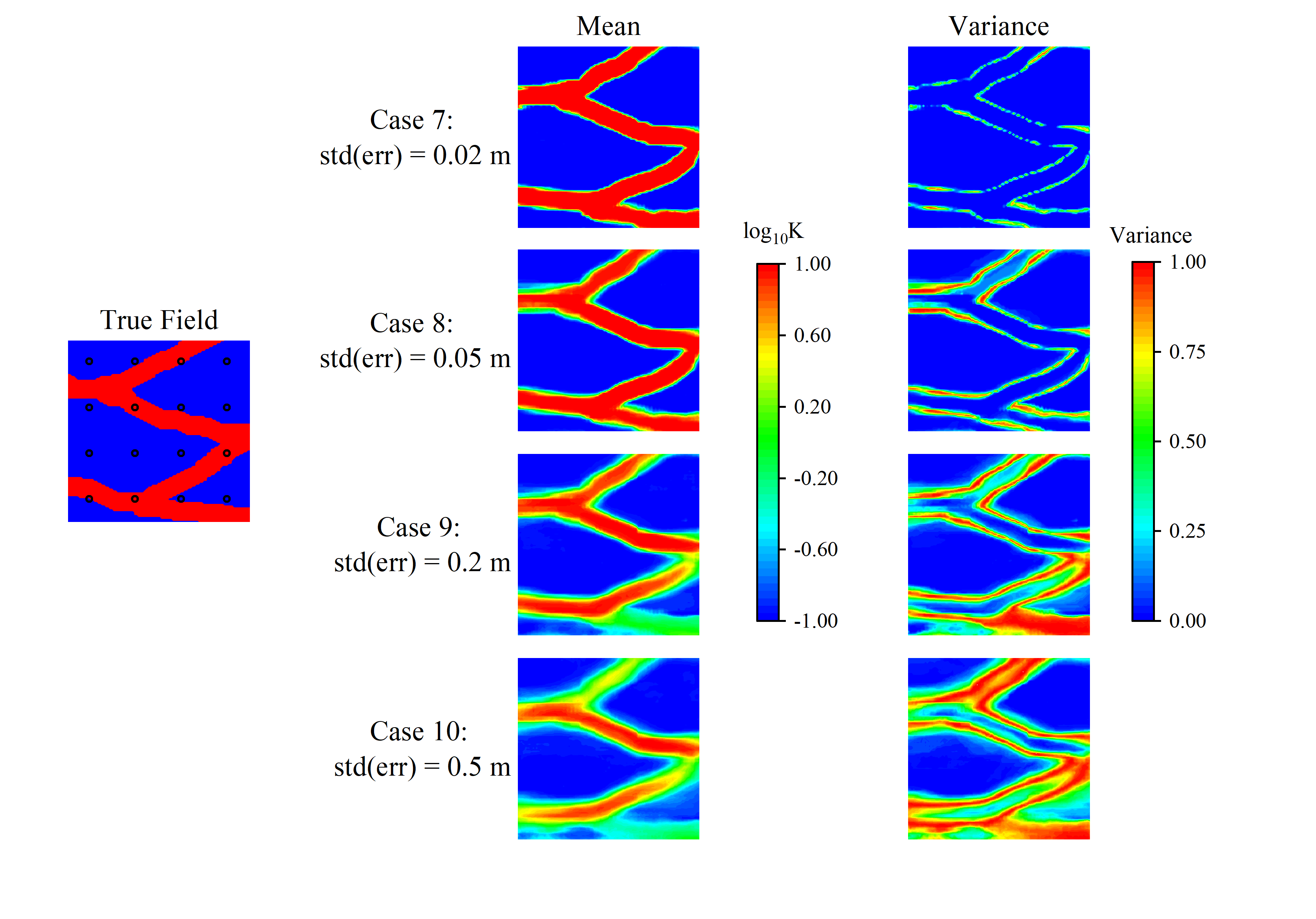}
\caption{Mean estimates and their posterior variance for different error levels of 0.02, 0.05, 0.2, and 0.5 m, respectively. The true field is shown on the left, and the black circles indicate well locations.
}
\label{fig:MeanVar_Error}
\end{figure}

In this subsection, we investigate the performance of the proposed approach under different error levels. As shown in Figure \ref{fig:MeanVar_Error}, four different error cases were considered in which Gaussian error levels of 0.02, 0.05, 0.2, and 0.5 m are added to the simulated data. The true field and observation layout are also shown in Figure \ref{fig:MeanVar_Error}. The observed data were obtained from 16 well HT at a pumping rate of 50 m$^3$/d. 

\begin{figure}[htbp]
\centering
\includegraphics[scale=0.2, angle=0]{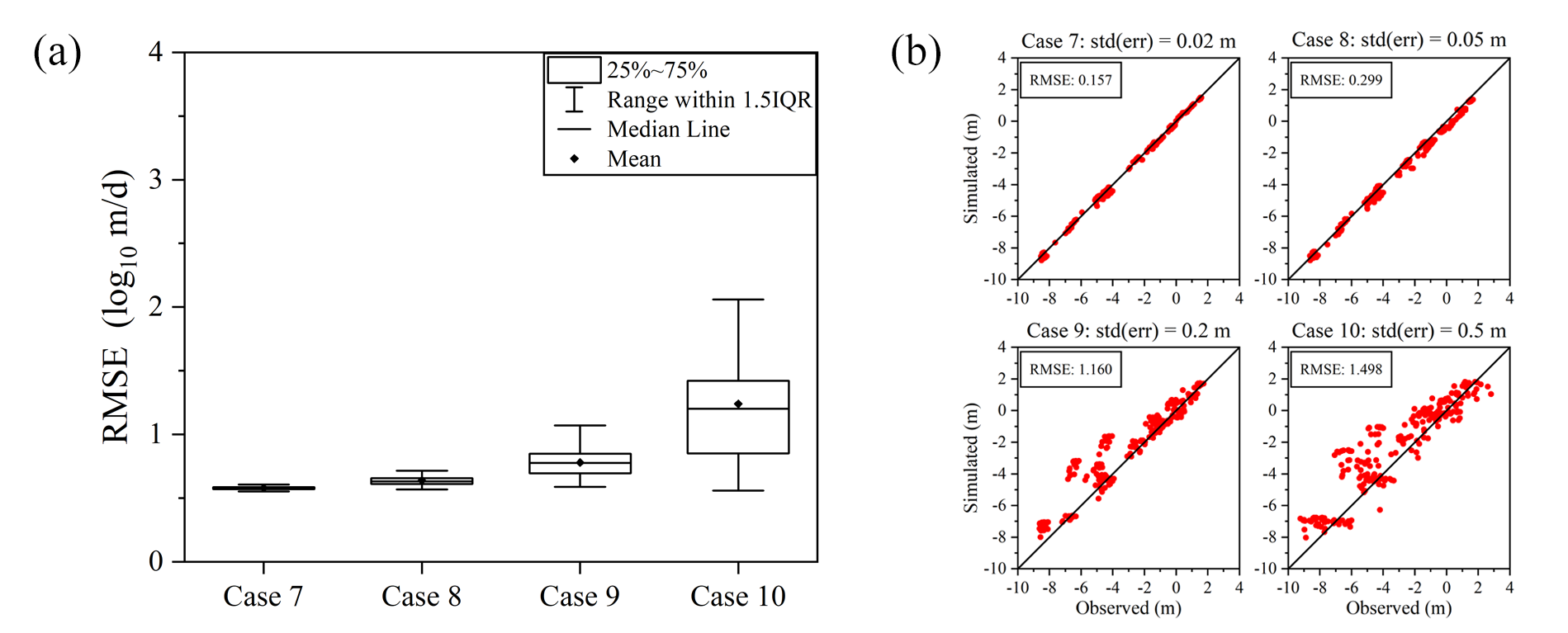}
\caption{(a) Boxplots of hydraulic conductivity RMSE (log$_{10}\ $m/d) for different error levels. (b) simulated hydraulic heads vs. observed hydraulic heads for different noise levels. The black lines are $45^{\circ}$ lines.
}
\label{fig:Fit_Error}
\end{figure}

Figure~\ref{fig:MeanVar_Error} shows the mean estimates and their estimation variance with different error levels. It is observed that our proposed approach can delineate the subsurface features accurately while the estimated hydraulic conductivity fields become blurred around the channel boundaries with larger uncertainty as the noise level increases. Even with the largest error level of 0.5 m, the mean estimate correctly covers the extent of the true high permeable channels. The boxplots in Figure~\ref{fig:Fit_Error} (a) show the RMSE values of estimated hydraulic conductivity for each case. As the error level increases, the RMSE value tends to increase and the RMSE range becomes wider as evidenced in Figure~\ref{fig:MeanVar_Error}. Figure~\ref{fig:Fit_Error} (b) presents the fitting errors of different error levels showing that the estimation is robust to the measurement error level up to 0.05 m in this case.  

% 0.02 std and 0.05 std are much smaller than the results of 0.2 std and 0.5 std. As the noise level increases, the observed data are less informative, and some data may be invalid.

\section{Discussion}
\label{sec:discussion}

\subsection{Non-Gaussianity and Dimension Reduction}
The proposed inverse modeling workflow by coupling WGAN-GP with ES-MDA showed satisfactory performance in both Gaussian and channelized test cases. For the Gaussian field, the trained generator plays a role in dimension reduction. The ES-MDA can be applied to the low-dimensional latent space instead of directly working on the high-dimensional hydraulic conductivity field. The 96 $\times$ 96 parameters of the conductivity field are reduced to 6 $\times$ 6 latent variables that need to be updated. For the channelized field, the trained generator plays another role in mapping a Gaussian field to a non-Gaussian field. The ES-MDA only works well for Gaussian distributed variables because of the Gaussian prior assumption, thus is not suitable for the non-Gaussian channelized field. However, the ES-MDA can be applied to the Gaussian distributed latent space using the trained generator, then the generator can produce the non-Gaussian channelized field. This conversion process nicely solves the limitation of ES-MDA or other Gaussian prior-based approaches.

\subsection{Compared with Variational Approach}
\begin{figure}[htbp]
\centering
\includegraphics[scale=0.4, angle=0]{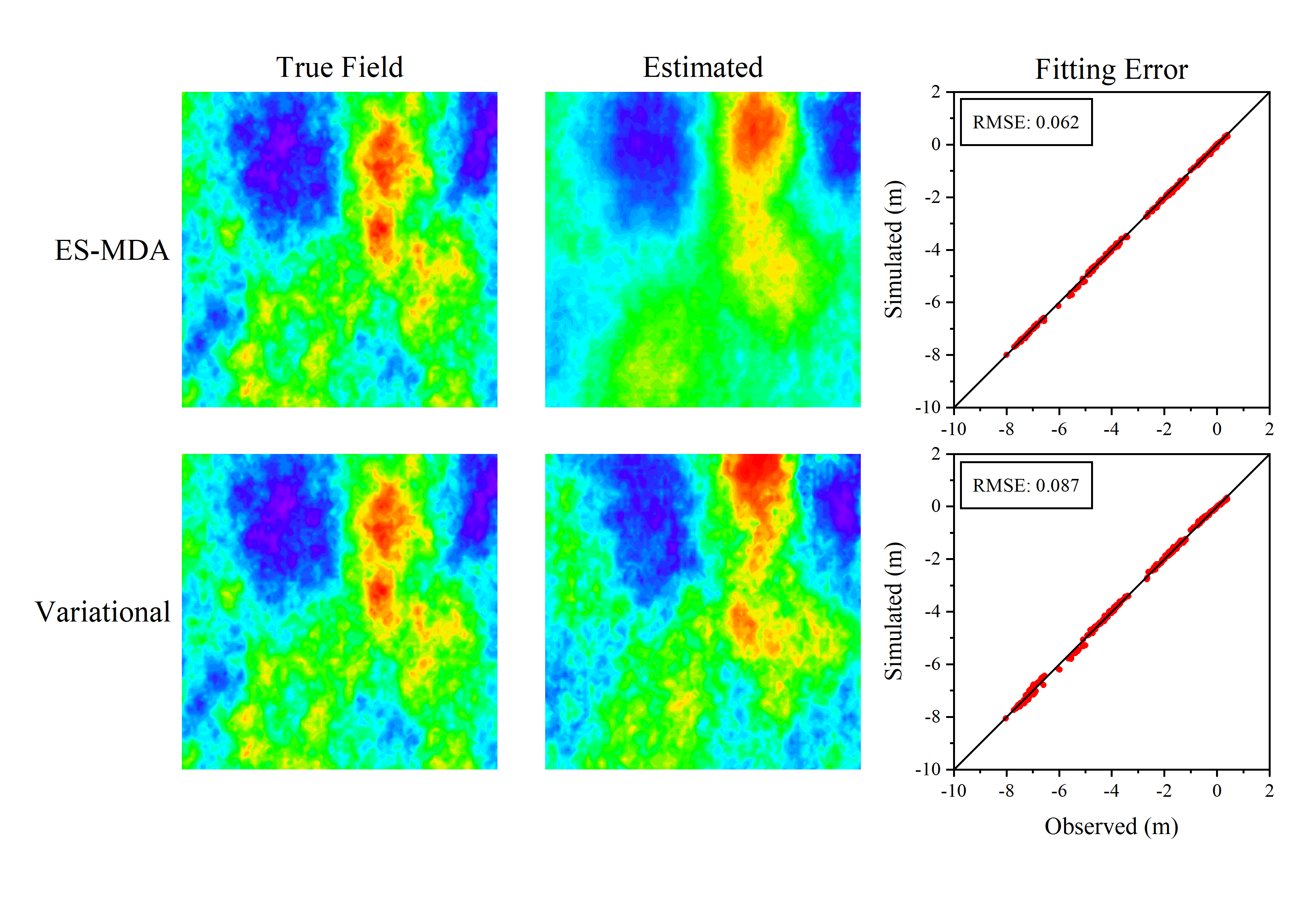}
\caption{Comparison of the results using ES-MDA and variational inversion for the Gaussian field}
\label{fig:Variational_Gaussian}
\end{figure}

\begin{figure}[htbp]
\centering
\includegraphics[scale=0.4, angle=0]{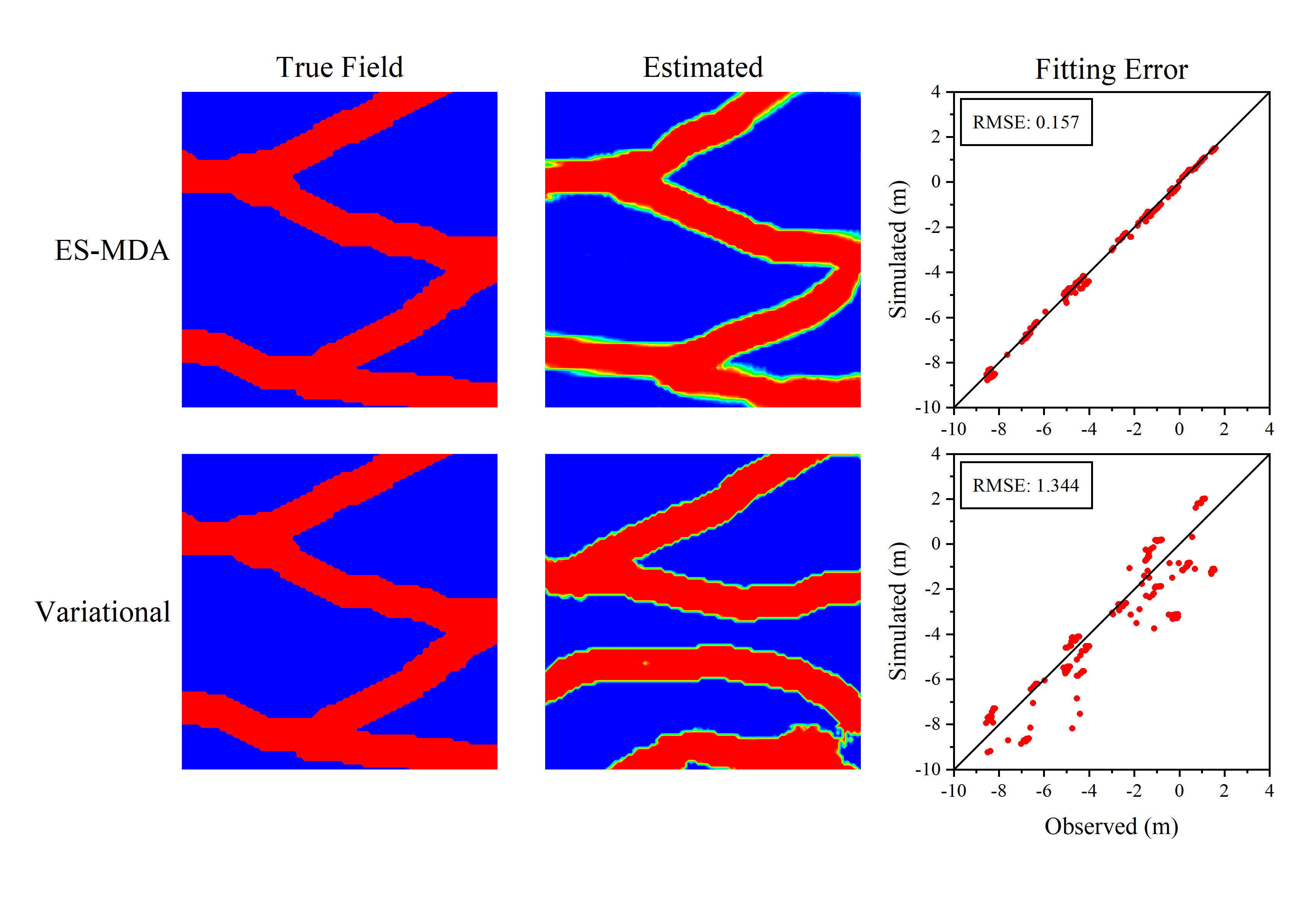}
\caption{Comparison of the results using ES-MDA and variational inversion for the channelized field}
\label{fig:Variational_Channel}
\end{figure}

Now, we investigate our choice of the ensemble-based approach, ES-MDA, for our main inversion algorithm over other inverse modeling methods. For this, we perform some experiments using variational inversion \cite{sasaki1958objective, sasaki1970some,forghani2022variational} with WGAN-GP prior to compare with the results using ES-MDA. The code used here for variational inversion implements a Gauss-Newton based optimization with a line search method as in \cite{forghani2022variational}. Figures \ref{fig:Variational_Gaussian} and \ref{fig:Variational_Channel} are the results for the Gaussian and channelized fields, respectively. Both ES-MDA and variational inversion perform well in the Gaussian case as observed in the previous researches \cite{lee2014large,forghani2022variational}. However, for the channelized aquifer problem, the variational inversion approach cannot recover the channel structures while ES-MDA can still achieve a good result. 

\subsection{Regularity of Latent Space}
In order to investigate why the ES-MDA method performs better in the channelized case, the objective function $L$ was calculated to visualize the regularity, i.e., roughness, of the latent space: 

\begin{equation}
    \label{eq:obj}
    L = \frac{1}{2}(\mathbf{z}_{ref}-\mathbf{z}_{pred})^{T} \mathbf{C}_{Z}^{-1}(\mathbf{z}_{ref}-\mathbf{z}_{pred}) + \frac{1}{2}(\mathbf{d}_{obs}-\mathbf{d}_{pred})^{T} \mathbf{C}_{D}^{-1}(\mathbf{d}_{obs}-\mathbf{d}_{pred})
\end{equation}
where $\mathbf{z}_{ref}$ represents the reference latent values, which are the latent values corresponding to a realization that is close to the true field as in Figure \ref{fig:MeanVar_Gaussian} and Figure \ref{fig:MeanVar_Channel}. $\mathbf{z}_{pred}$ represents the predicted latent values for each iteration of the data assimilation or the inversion process. $\mathbf{C}_{Z}$ is the latent error matrix. By the construction of the latent variables following the Gaussian distribution $N(0,1)$, thus the diagonal entries of $\mathbf{C}_{Z}$ and $\mathbf{C}_{Z}^{-1}$ are 1. $\mathbf{d}_{obs}$ is the observed data, $\mathbf{d}_{pred}$ is the predicted data, and $\mathbf{C}_{D}$ is the observed error matrix. We assume the observed error follows the Gaussian distribution $N(0,0.02^{2})$, thus the diagonal entries of $\mathbf{C}_{D}^{-1}$ are $1/(0.02^{2})=2,500$. Both variational inversion and ES-MDA minimize the same objective function \ref{eq:obj}, however, their minimization approaches are different; the variational inversion directly applies the Gauss-Newton method \cite{nocedal2006numerical} to the objective function (Eq. \ref{eq:obj}) while the ensemble-based approach approximates the covariance matrices in the objective function by an ensemble or generated samples and then perform the optimization by minimizing their expected objective function value \cite{anderson2003local,lee2018riverine}. It is shown that the accurate covariance approximation from an ensemble requires a number of generated realizations \cite{johnstone2001distribution,elkaroui2008spectrum}, which is generally regarded as a limitation of the ensemble-based inversion since it makes the objective function space smooth and leads to a suboptimal solution \cite{lee2018riverine}. However, for the applications presented in this study, this ensemble-based approximation error actually regularizes the problem in a beneficial way and provides better solutions. 

\begin{figure}[htbp]
\centering
\includegraphics[scale=0.18, angle=0]{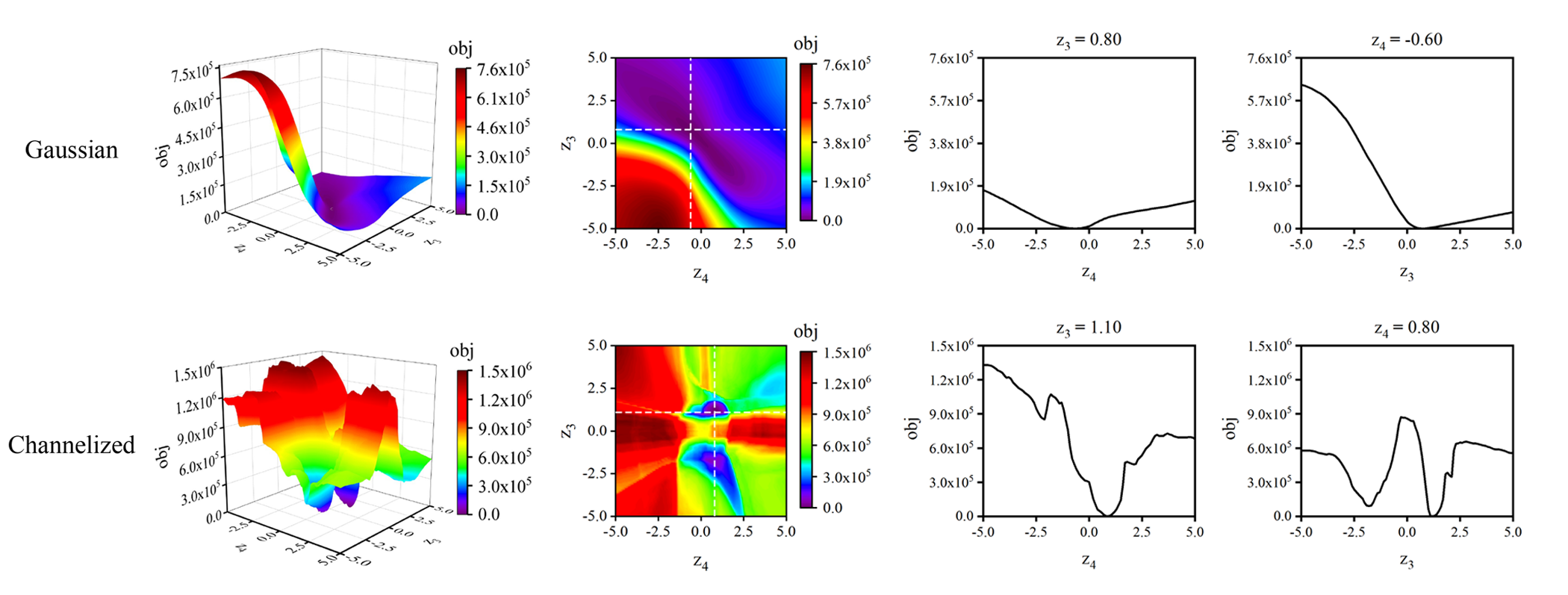}
\caption{Objective functions with respect to $\mathbf{z}_{3}$ and $\mathbf{z}_{4}$. The first row is the Gaussian case, and the second row is the channelized case. The first and second columns are the 3D and 2D plots of the objective functions. The third and fourth columns are the plots near the reference values of $\mathbf{z}_{3}$ and $\mathbf{z}_{4}$. The white lines in the second column indicate the locations of the plots.  
}
\label{fig:Obj_Fun}
\end{figure}

Figure \ref{fig:Obj_Fun} shows an example of objective functions with respect to $\mathbf{z}_{3}$ and $\mathbf{z}_{4}$ for the Gaussian and channelized cases, where $\mathbf{z}_{i}$ is the $i$-th latent space variable. We chose third and fourth latent variables $\mathbf{z}_{3}$ and $\mathbf{z}_{4}$ for effective illustration but the patterns of the 9 (3 $\times$ 3) latent variables are similar. $\mathbf{z}_{3}$ and $\mathbf{z}_{4}$ are sampled from [-5,5] with an interval of 0.1. The channelized space is pretty rough due to the nonlinear, aggressive (96 $\times$ 96 to 3 $\times$ 3) dimension reduction, while the Gaussian space is much smoother and suitable for the Gauss-Newton type methods as expected \cite{lee2014large}. The third and fourth columns in Figure \ref{fig:Obj_Fun} are the 1D plots near the reference values of $\mathbf{z}_{3}$ and $\mathbf{z}_{4}$. For the Gaussian case, the minimum values are clearly shown in the plots. However, the channelized case has several local minima, which may hinder the algorithms from finding the global minima unless one carefully applies non-smooth function optimization techniques \cite{hiriart1993convex,nocedal2006numerical}.         

\begin{figure}[htbp]
\centering
\includegraphics[scale=0.3, angle=0]{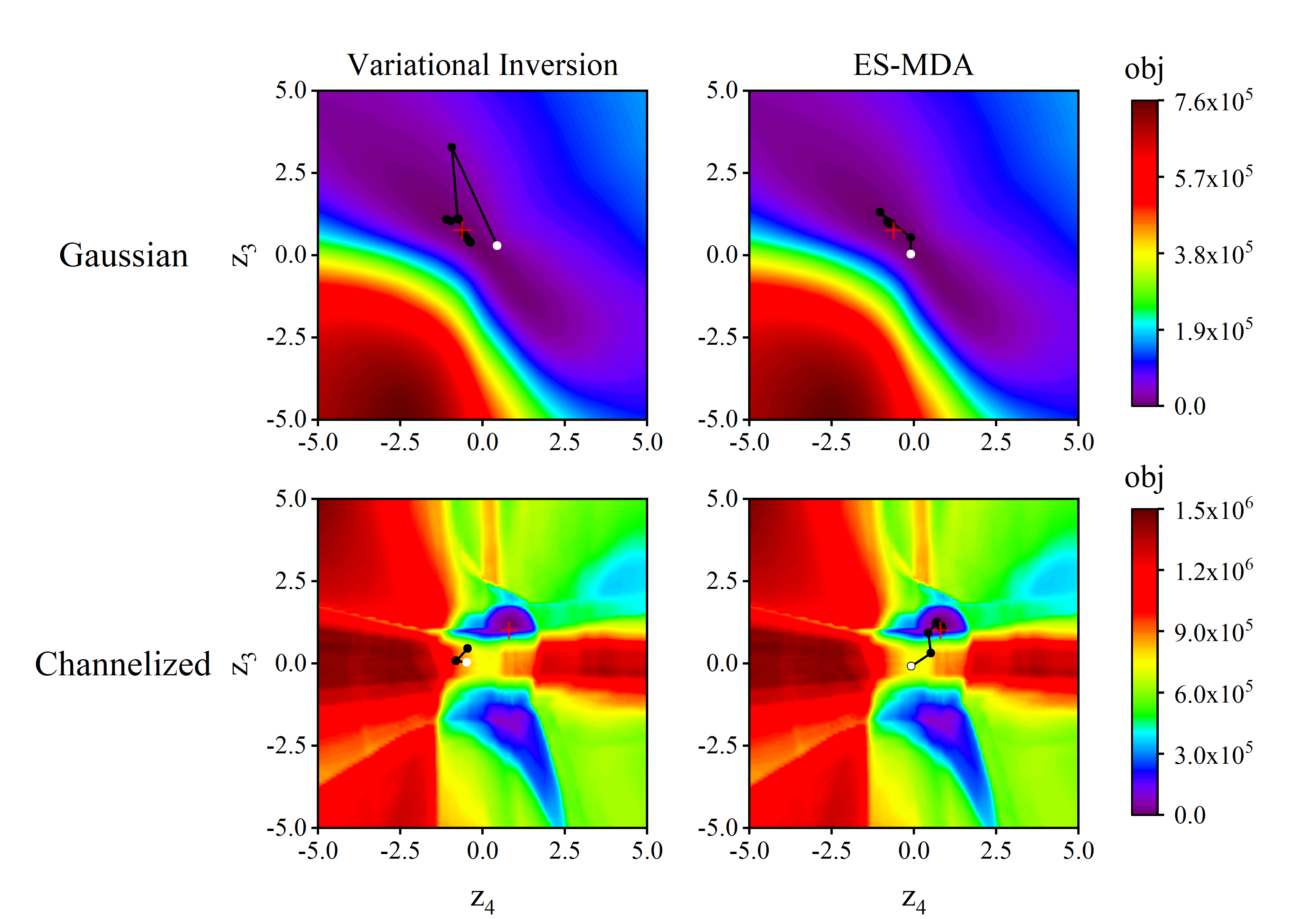}
\caption{Update paths with respect to $\mathbf{z}_{3}$ and $\mathbf{z}_{4}$. The first column shows the results using variational inversion for the Gaussian and channelized cases. The second column shows the results using ES-MDA, only the mean of 200 realizations is shown here. Red crosses indicate reference values, white dots indicate starting points, and black dots represent predicted values.      
}
\label{fig:Update_Path}
\end{figure}

Figure \ref{fig:Update_Path} shows the Gauss-Newton update paths of variational inversion and ES-MDA. For the Gaussian case, both methods can reach some points close to the reference (i.e., optimal) values. For the channelized case, however, variational inversion is more likely to get stuck in some local minima, while ES-MDA can identify the correct optimization paths to the global minimum through an average gradient for the ensemble \cite{anderson2003local}. Variational inversion here implemented the Gauss-Newton method with simple line search \cite{forghani2022variational} and Levenberg–Marquardt algorithms \cite{nowak2004modified} but still could not handle the non-linearity. For non-smooth functions as in the channelized case, it is easy to converge into local minima as in Figure \ref{fig:Update_Path}. ES-MDA updates multiple realizations at the same time and the expected Gauss-Newton update step at each iteration is determined by the average of individual update in each realization. Such strategy can select feasible minimization directions and effectively make the objective function surface smooth for better convergence at the cost of more function evaluations.

\subsection{Fractured Field Application}
\begin{figure}[htbp]
\centering
\includegraphics[scale=0.4, angle=0]{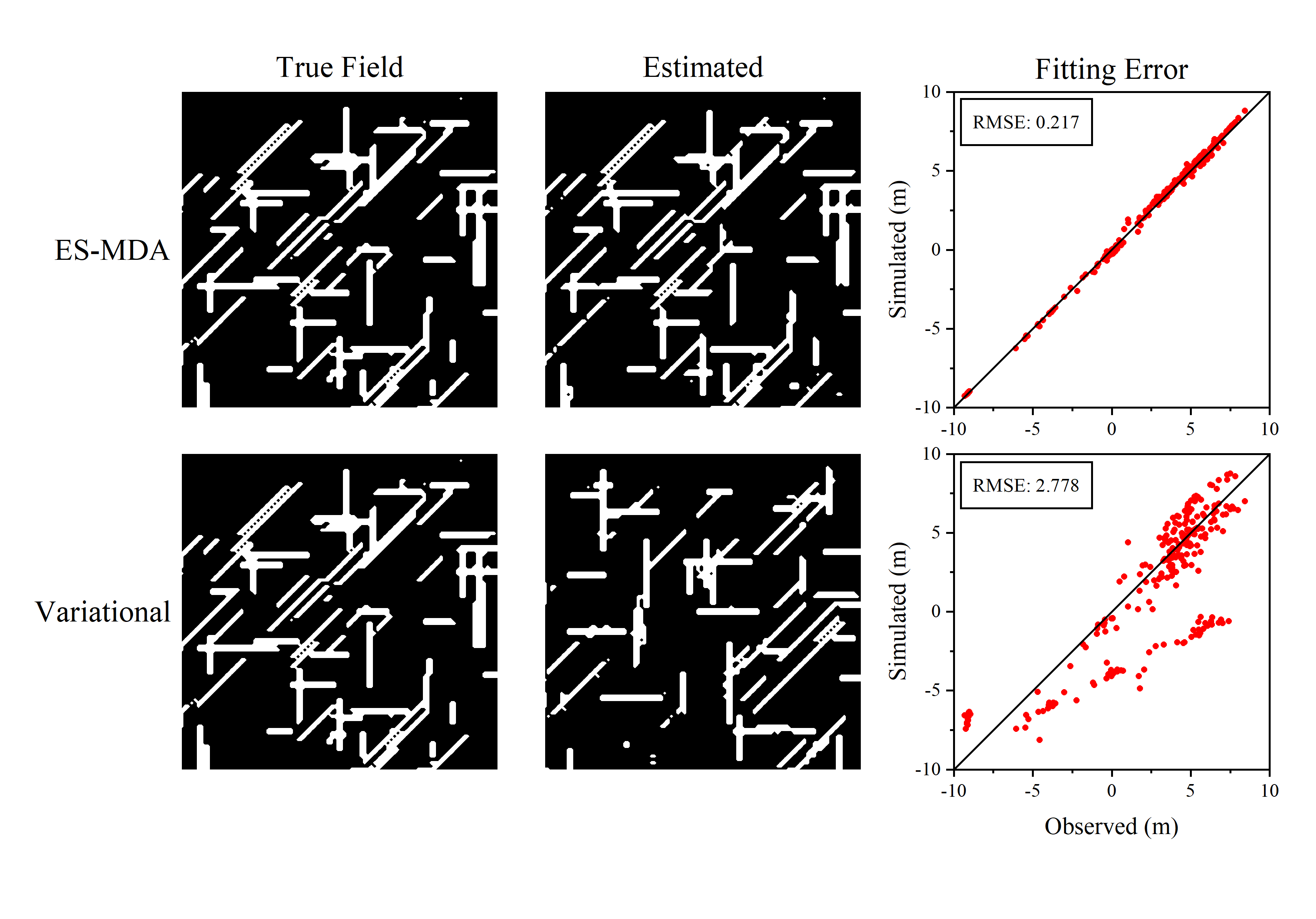}
\caption{Comparison of the results using ES-MDA and variational inversion for the fractured field}
\label{fig:Fractures}
\end{figure}

In this section, a fractured field was tested to demonstrate the application of the proposed approach for another difficult inverse problem: discrete fractured media or local-scale high-permeable inclusions. 80,000 images with randomly generated fractures were used for training. The fractures have 3 orientations: $0^{\circ}$, $45^{\circ}$, and $90^{\circ}$. The fractures are 10 to 20 pixels long and 2 pixels wide. The network architecture is the same as in the channelized case (Table~\ref{tab:WGAN_Channel}). To show the performance comparison clearly, we select an image generated by the trained generator as the true field for experiments. The model settings including boundary conditions and well configurations are the same as shown in Figure~\ref{fig:Model}, and the estimation results are shown in Figure~\ref{fig:Fractures}. To visualize the estimated fractures clearer, we applied a threshold of 0.2 for final binary image production. All the fractures have a log-conductivity of 1 ($\log_{10}$ in m/d as white fractures in Figure~\ref{fig:Fractures}), and the background matrix has a log-conductivity of -1 ($\log_{10}$ in m/d as background black domain in Figure~\ref{fig:Fractures}). As expected, the ES-MDA method performs better than the variational approach by reconstructing most of fracture features and reproducing the observations with a much smaller RMSE value of 0.217 m. 

\subsection{Extrapolation Capability of WGAN-GP/ES-MDA Inversion Framework}
\begin{figure}[htbp]
\centering
\includegraphics[scale=0.4, angle=0]{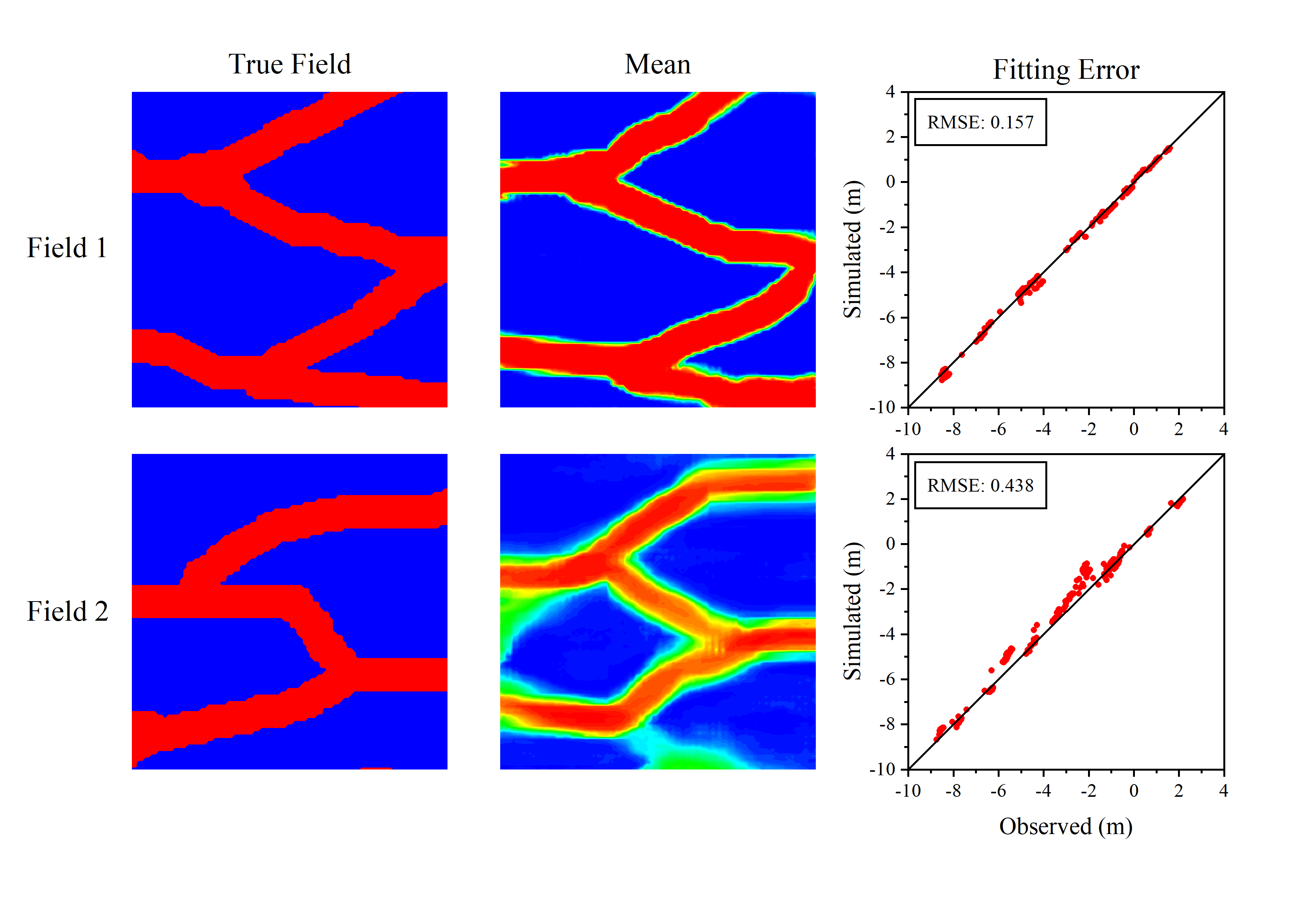}
\caption{Comparison of different true fields. }
\label{fig:Different_Ref}
\end{figure}

Lastly, we test the estimation accuracy of our inversion framework with a true field beyond the generation capability of WGAN-GP. For this, we intentionally select one of the test data as the true field that is most far away from realizations generated from WGAN-GP and perform the inversion with the observations from the true field as shown in Field 2 of Figure \ref{fig:Different_Ref}). For the comparison purpose, we plot the result of Case 7 from the previous section for a true $K$ field that is randomly chosen from the test data set (Field 1 in Figure \ref{fig:Different_Ref}). As shown in Figure \ref{fig:Different_Ref}, the results of the Field 2 case are worse than those from the Field 1 case simply because the trained generative model does not have the power to reproduce Field 2, especially connected structures near the domain boundaries. As a result, the mean estimate of the Field 2 estimation is more blurry with the channel shapes slightly shifted from the true field. Field 1 has a fitting error of RMSE 0.157 m and Field 2 has a much larger fitting error of RMSE 0.438 m. Still, the proposed framework carries out the right inversion task within its generation capability and estimates the field as close to the true field as possible. This indicates a potentially superior extrapolation capability of our proposed work especially when the deep generative model misses important subsurface features in the aquifer.   

\section{Conclusion}
\label{sec:conclusion}
In this work, a data assimilation/inversion workflow coupling Wasserstein generative adversarial network with gradient penalty (WGAN-GP) with ensemble smoother with multiple data assimilation (ES-MDA) is proposed for accurate and scalable subsurface characterization. WGAN-GP addresses the limitations of adversarial generative models such as mode collapse and ensures an accurate representation of underlying complex spatial fields. ES-MDA is intended to search for an optimal inverse solution without introducing user-controlled optimization parameters.

The trained generators of WGAN-GP were used to reparameterize high-dimensional conductivity fields with low-dimensional latent spaces and ES-MDA was then used to update the latent variables. Through this coupling, the non-Gaussian fields can be mapped to a Gaussian distributed space via WGAN-GP, which allows ES-MDA to work properly with its internal Gaussian assumption. We applied the proposed method to several test cases that aim to identify unknown Gaussian, channelized, and fractured hydraulic conductivity fields from sparse data sets. The numerical experiments show that the 96 $\times$ 96 parameters of the conductivity fields can be reduced to 6 $\times$ 6 (Gaussian case) and 3 $\times$ 3 (channelized and fractured cases) latent variables. The results illustrate: (1) coupling WGAN-GP with ES-MDA can reconstruct the main features of the Gaussian field as expected and accurately estimate the complex small-scale features in the channelized and fractured fields; (2) the proposed method can accommodate more information and accordingly improve the estimation results with lower uncertainty, for example, by installing more observation wells, cross-well pumping tests such as hydraulic tomography with fewer wells, and/or geophysical sensing data; (3) the proposed approach is robust to the data contaminated with high errors (e.g., std(err) = 0.2 std and 0.5 m) and the mean estimate can still identify the main underlying complex features with reasonable uncertainty quantification. 

One of the main contributions of this paper is to justify the use of ensemble-based approaches when using deep generative models. Moreover, the proposed approach outperforms variational inversion for minimizing non-smooth, multimodal objective functions as shown in the channelized and fractured cases. It is observed that multiple realizations examined in the proposed method lead to easier objective function minimization with prudent Gauss-Newton steps, thus the proposed method is less likely to get stuck in local minima and can converge to the close-to-optimal inverse solution with better estimation accuracy. However, the variational approach can reduce the number of forward model runs up to the latent variable dimension, for example, about 10 forward model runs at each iteration. Additional development and modification are needed for the variational inversion approach that ensures local and global convergence using robust techniques such as Trust-region methods \cite{boyd2004convex,nocedal2006numerical}.  

The proposed approach is also computationally efficient, and it is easy to apply parallelization. The data assimilation process takes around 7 minutes with 200 realizations and 8 iterations on a workstation with 64G RAM Intel(R) Core(TM) i9-9900X CPU @ 3.50GHz. The coupled WGAN-GP/ES-MDA python code used in this paper can be found in the \url{https://github.com/jichao1/WGAN-GP.git}.

\acknowledgments
This work is supported by the US Department of Energy Office of Fossil Energy and Carbon Management project - Science-Informed Machine Learning to Accelerate Real-Time Decisions in Subsurface Applications (SMART) initiative. Sandia National Laboratories is a multi-mission laboratory managed and operated by National Technology \& Engineering Solutions of Sandia, LLC (NTESS), a wholly owned subsidiary of Honeywell International Inc., for the U.S. Department of Energy’s National Nuclear Security Administration (DOE/NNSA) under contract DE-NA0003525. This written work is authored by an employee of NTESS. The employee, not NTESS, owns the right, title, and interest in and to the written work and is responsible for its contents. Any subjective views or opinions that might be expressed in the written work do not necessarily represent the views of the U.S. Government. The publisher acknowledges that the U.S. Government retains a non-exclusive, paid-up, irrevocable, world-wide license to publish or reproduce the published form of this written work or allow others to do so, for U.S. Government purposes. The DOE will provide public access to results of federally sponsored research in accordance with the DOE Public Access Plan.

\bibliography{references}

\end{document}